\documentclass{article} 
\usepackage[accepted]{icml2024}

\usepackage{amsmath,amsfonts,bm}

\def\eqref#1{equation~\ref{#1}}

\def\1{\bm{1}}

\DeclareMathAlphabet{\mathsfit}{\encodingdefault}{\sfdefault}{m}{sl}
\SetMathAlphabet{\mathsfit}{bold}{\encodingdefault}{\sfdefault}{bx}{n}

\usepackage[hidelinks]{hyperref}
\usepackage{url}
\usepackage{pdfpages}
\usepackage{caption}
\usepackage{subcaption}
\usepackage{algorithm} 
\usepackage{pgfplots}
\usepackage{tikz}
\usepackage{listings}
\usepackage{pdfpages}
\usepackage{booktabs}
\usepackage{todonotes}
\usepackage{multirow}
\usepackage[symbol*]{footmisc}
\usetikzlibrary{shapes.geometric, arrows}
\usetikzlibrary{dsp, chains}

\pgfplotsset{compat=1.18}

\icmltitlerunning{Fourier-Based Neural Operators on Arbitrary Domains}

\begin{document}

\twocolumn[
\icmltitle{Beyond Regular Grids: Fourier-Based Neural Operators on Arbitrary Domains}



\icmlsetsymbol{equal}{*}

\begin{icmlauthorlist}
\icmlauthor{Levi Lingsch}{ethmath}
\icmlauthor{Mike Y. Michelis}{ethai,srl}
\icmlauthor{Emmanuel de B\'ezenac}{ethmath}
\icmlauthor{Sirani M. Perera}{erau}
\icmlauthor{Robert K. Katzschmann}{ethai,srl}
\icmlauthor{Siddhartha Mishra}{ethmath,ethai}
\end{icmlauthorlist}

\icmlaffiliation{ethmath}{Seminar for Applied Mathematics, ETH Zurich,  Switzerland}
\icmlaffiliation{ethai}{ETH AI Center, ETH Zurich, Switzerland}
\icmlaffiliation{srl}{Soft Robotics Lab, ETH Zurich, Switzerland}
\icmlaffiliation{erau}{Department of Mathematics, Embry-Riddle Aeronautical University, Daytona Beach, FL, USA}
\icmlcorrespondingauthor{Levi Lingsch}{llingsch@student.ethz.ch}

\icmlkeywords{Operator Learning, Fast Fourier Transforms, }

\vskip 0.3in
]
\printAffiliationsAndNotice{} 


\begin{abstract}

The computational efficiency of many neural operators, widely used for learning solutions of PDEs, relies on the fast Fourier transform (FFT) for performing spectral computations. As the FFT is limited to equispaced (rectangular) grids, this limits the efficiency of such neural operators when applied to problems where the input and output functions need to be processed on general non-equispaced point distributions. Leveraging the observation that a limited set of Fourier (Spectral) modes suffice to provide the required expressivity of a neural operator, we propose a simple method, based on the efficient direct evaluation of the underlying spectral transformation, to extend neural operators to arbitrary domains. An efficient implementation\footnote{Source code available on GitHub: \url{https://github.com/camlab-ethz/DSE-for-NeuralOperators}}  of such \emph{direct spectral evaluations} is coupled with existing neural operator models to allow the processing of data on arbitrary non-equispaced distributions of points. With extensive empirical evaluation, we demonstrate that the proposed method allows us to extend neural operators to arbitrary point distributions with significant gains in training speed over baselines while retaining or improving the accuracy of Fourier neural operators (FNOs) and related neural operators.

\end{abstract}

\vspace{-1em}

\section{Introduction}
\label{sec:introduction}
Partial Differential Equations (PDEs) are extensively used to mathematically model interesting phenomena in science and engineering \cite{Evansbook}. As closed-form or analytical solutions to solve PDEs are not available or practical, traditional numerical methods such as finite difference, finite element, and spectral methods \cite{NAbook} are used to solve PDEs. Despite their tremendous success, the prohibitively high computational cost of these methods makes them infeasible for a variety of contexts in PDEs ranging from high-dimensional problems to the so-called \emph{many query} scenarios \cite{KarRev}. This high computational cost also provides the rationale for the development of alternative \emph{data driven} methods for the fast and accurate simulation of PDEs. Hence, a wide variety of machine learning algorithms have been proposed recently in this context. These include physics-informed neural networks (PINNs) \cite{KAR2}, MLPs, and CNNs for simulating parametric PDEs \cite{ZhuZab1,LMR1,ISMO,Wandel_Fluids} as well as graph-based algorithms \cite{Sanchez_Graphs, Pfaff_GNN,Well,ERM1}, to name a few.  

However, as solutions of PDEs are expressed in terms of the so-called \emph{solution operators}, which map input functions (initial and boundary data, coefficients, source terms) to the PDE solution, \emph{operator learning}, i.e., learning the underlying operators from data, has emerged as a dominant framework for applying machine learning to PDEs. Existing operator learning algorithms include, but are not limited to, operator networks \cite{chenchen}, DeepONets \cite{Lu_DeepONet,donet1,donet2}, attention-based methods such as \cite{loca,cao1,vidon}, and neural operators \cite{NO,LNO,MNO,CNO}. More recently, transformer-based approach for learning operators have been explored \cite{wu2023solving, li2023scalable}, yet operator learning in the spectral domain retains an advantage \cite{hao2024dpot}.  

\begin{figure}
    \centering
    \includegraphics[width=0.49\textwidth,trim=80 390 460 60, clip]{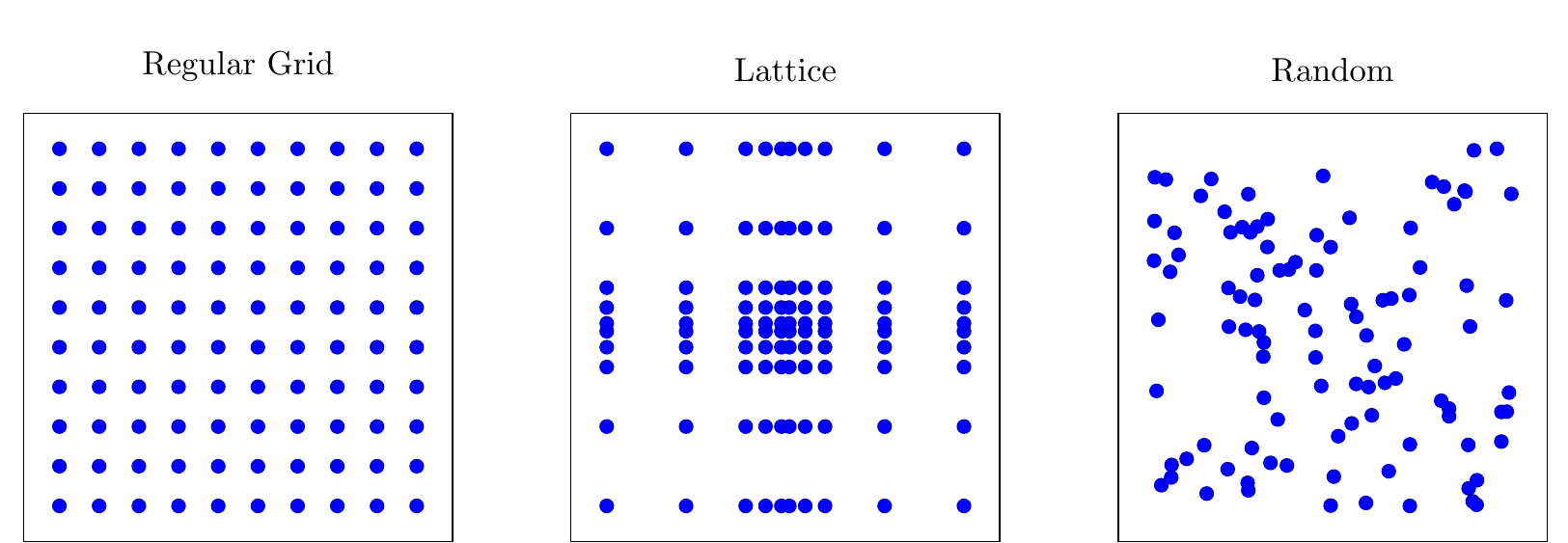}
        \caption{Distributions discussed in this paper. The vanilla FNO is restricted to the regular grid, but may be applied to a general lattice distribution, random distribution, or structured point cloud via the approach outlined in Section 2.}
    \label{fig:Point Distributions}
    \vspace{-0.5em}
\end{figure}

Within this large class of operator learning algorithms, Neural Operators based on \emph{non-local} spectral transformations, such as the Fourier Neural operator (FNO) \cite{Li_FNO} and its variants \cite{FNO1,FNO2} have gained much traction and 
are widely applied. Apart from favorable theoretical approximation properties \cite{Kovachki_FNOBounds,LMHM1}, FNOs are attractive due to their expressivity, simplicity, and computational efficiency. 
A key element in the computational efficiency of the FNO and its variants lies in the fact that its underlying convolution operation is efficiently carried out in Fourier space with the \emph{fast Fourier transform} (FFT) algorithm. It is well-known that the FFT is only (log-)linear in computational complexity with respect to the number of points at which the underlying input functions are sampled. However, this computational efficiency comes at a cost as the recursive structure of the FFT limits its applications to inputs sampled on the so-called \emph{Regular} or \emph{equispaced Cartesian (Rectangular) grids}, see Figure~\ref{fig:Point Distributions}~left for an illustration. This is a major limitation in practice. In real-world applications, where information on the input and output signals is measured by sensors, it is not always possible to place sensors only on an equispaced grid. Similarly, when data is obtained through numerical simulations, often it is essential to discretize PDEs on irregular grids, such as those adapted to capture relevant spatially localized features of the underlying PDE solution or on unstructured grids that fit the complex geometry of the underlying domain. See Figure~\ref{fig:Point Distributions} for examples of such non-equispaced distributions of sample points or \emph{point clouds}. 

Given this context, it is \emph{imperative to design neural operators that can process input and output functions sampled on arbitrary point distributions.} Consequently, several methods have been proposed in the literature to address this limitation of FNOs and modify/enhance it to handle data on non-equispaced points. A straightforward fix would be to interpolate data from non-equispaced point distributions to equispaced grids. However, as shown in \cite{Li_GeoFNO}, the resulting procedure can be computationally expensive and/or inaccurate. \citet{Li_GeoFNO} propose a geometry-aware FNO (Geo-FNO) that appends a neural network to the FNO to learn a deformation from the underlying domain to a regular grid. Then, the standard FFT can be applied to the latent space of equispaced grid points. This learned diffeomorphism corresponds to an adaptive moving mesh \cite{Huang_AdaptiveMesh}. Factorized-FNO (F-FNO) builds upon the Geo-FNO, introducing an additional bias term in the Fourier layer and performing the Fourier transform over each dimension separately \cite{tran2023factorized}. The non-equispaced Fourier PDE solver (NFS) uses a vision mixer \cite{Dosovitskiy_VisionMixer} to interpolate from a non-equispaced signal onto a regular grid, again applying the standard FNO subsequently \cite{Lin_NFS}. All these methods share the same design principle, i.e., given inputs on non-equispaced points, \emph{interpolate} or transform this data into a regular grid and then apply the FNO, leading to a natural question: \emph{Is there an alternative approach where truncated spectral transformations, such as the discrete Fourier transform (DFT) inside FNO, can be performed efficiently on arbitrary point distributions? } 

The main goal of this paper is to propose such an alternative approach. Our starting point is the recent observation, both theoretical as well as empirical, in \cite{LMHM1,lanthaler2023nonlocal} and references therein, that in practice, only a relatively small (fixed) number of Fourier modes suffice to provide the required expressivity of FNOs and the superior performance of FNO can be attributed to other factors such as residual connections, high-dimensional lifting operators and nonlinear activations. Moreover, the number of \emph{effective} Fourier modes is significantly smaller than (and independent of) the number of points at which the input and output functions are sampled. This motivates us to design a simple method based on an \emph{efficient direct evaluation of truncated spectral transformations within neural operators to extend them to arbitrary point clouds.} More concretely, our contributions to this paper are



\begin{itemize}
\item We present a simple, efficient method to formulate truncated \emph{spectral transformations}, such as the Fourier transform or spherical harmonic transform, on arbitrary point distributions. A \emph{novel} PyTorch implementation of these direct spectral evaluations is also presented.
\item We \emph{replace} the standard truncated spectral transformations that underpin many widely-used neural operators by the presented direct spectral evaluation, to obtain neural operators that can efficiently handle input and output data on arbitrary sample points over domains with arbitrary geometries. 
\item We present a suite of extensive numerical experiments to demonstrate that a variety of neural operators, based on truncated spectral transformations computed using the proposed direct approach, outperform baselines in terms of both accuracy and efficiency (training speed) in various scenarios involving input/output data on arbitrary point distributions. 
\end{itemize}

\begin{figure*}
    \centering
    \includegraphics[width=1.0\textwidth, trim=70 330 360 70, clip]{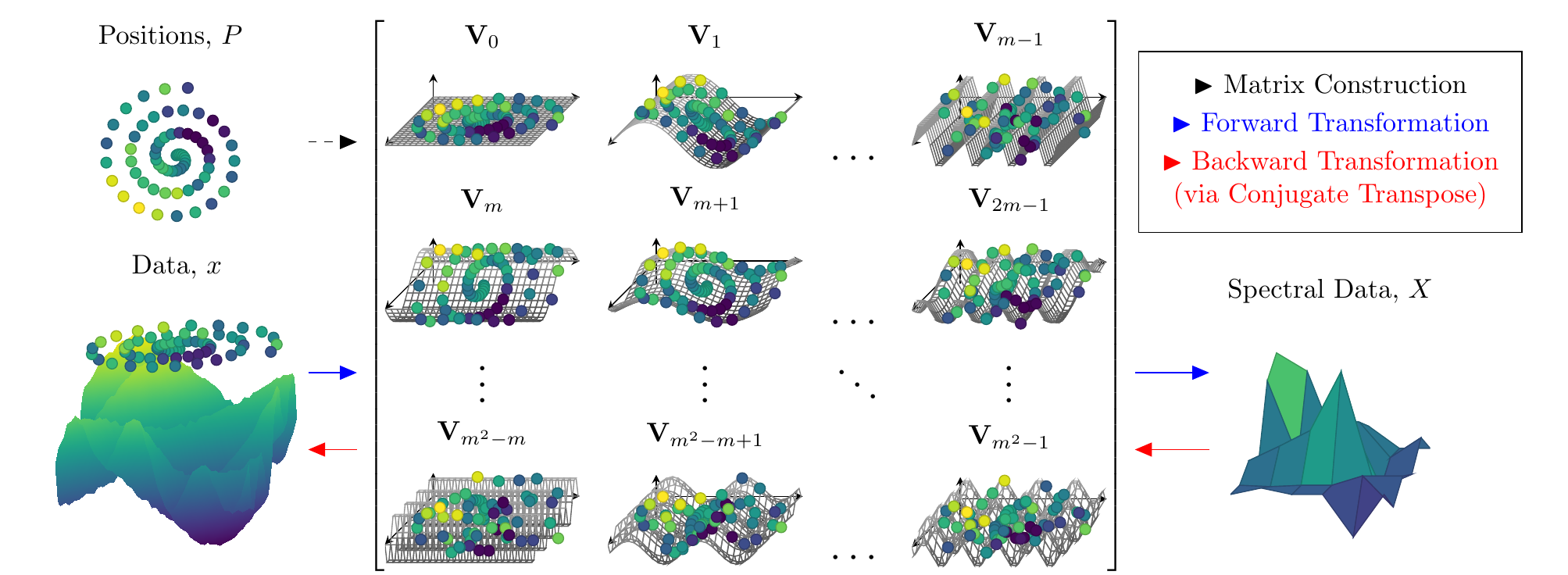}
    \caption{Illustration of the truncated Fourier transform on a point cloud. We construct a matrix using the values of a set of Fourier basis functions sampled at the positions of the sample points. The transformation between spatial and spectral domains is directly evaluated via a matrix-vector product, where the vector contains the training data. The subscript $\mathbf{V_j}$ refers to the $j^{th}$ row of the matrix from Equation (\ref{eq:flattened structured matrix}).}
    \label{fig:main figure}
    \vspace{-0.5em}
\end{figure*}

\section{Methods}
\label{sec:Vandermonde_Transform}

Transforms, such as the Fourier transform and spherical harmonics, form the foundations of many widely-used neural operators such as FNO \citep{Li_FNO}, UFNO \citep{UFNO}, F-FNO \citep{tran2023factorized} and SFNO \citep{SFNO}. In this section, we describe the calculation of discrete forms of these transforms over nonuniform data and the construction of suitable matrices to compute these transforms efficiently within neural operators. To this end, we start with a short description of this matrix construction in the 1D case. 


\textbf{Forward Transformations.} Any discretization of the Fourier transform will transform a sequence of $N$ complex numbers $\mathbf{x} = [x_0, \hdots, x_{N-1}]^T$ into another sequence of complex numbers $\mathbf{X} = [X_0, \hdots, X_{N-1}]^T$ via 
\begin{equation}
\label{eq:1}
    X_{k}=\sum_{n=0}^{N-1}x_{n}e^{-2 \pi i p_{n}f_{k}},\quad0\leq k\leq N-1.
\end{equation}
Here, $p_0, \hdots, p_{N-1} \in [0,1]$ are sample point positions, $f_0, \hdots, f_{N-1}\in[0,N]$ are frequencies, and $i=\sqrt{-1}$. In this work, we focus on using nonuniform sample points with uniform (integer) frequencies; this is commonly referred to the type II NUDFT \citep{NUDFT_greengard}. 

Modern GPUs and machine learning libraries are optimized for handling matrices. We will construct a matrix of the form,
\begin{equation}
    \mathbf{V}_{j,k} = \frac{1}{\sqrt{N}}\left[ e^{-2 \pi i \left( {j} p_k\right)} \right]_{j,k=0}^{m-1, N-1},
    \label{eq:V_1d}
\end{equation}
where $m$ the number of modes, truncated such that $m << N$. Then, we can computationally realize the transform (\ref{eq:1}) by a matrix-vector product
\begin{equation}
    \mathbf{X} = \mathbf{V} \mathbf{x},
    \label{eq:transform}
\end{equation}
relying on the optimized matrix operations provided by popular machine learning libraries, such as PyTorch. Furthermore, support for batched matrix multiplications is already present, allowing this approach to be used for batched data.

\paragraph{Extension to D-dimensional arbitrary point clouds.} The multidimensional NUDFT may be presented as 
\begin{equation}
X_{\boldsymbol{k}}=\sum_{\boldsymbol{n}=0}^{\boldsymbol{N}-1}x_{\boldsymbol{n}}e^{-2\pi i{\boldsymbol{p}}_{\boldsymbol{n}}\cdot \boldsymbol{f}_{\boldsymbol{k}}},
\end{equation}
transforming a $D$-dimensional array of complex numbers $x_{\boldsymbol{n}}$ into another complex $D$-dimensional array $X_{\boldsymbol{k}}$. In this case, $\boldsymbol{p_n} \in [0,1]^D$ are sample points, $[0,1]^D:=[0,1] \times [0,1] \times \cdots \times [0,1]$ ($D$ times),  $\boldsymbol{f_k}\in [0,N_1]\times [0,N_2]\times \cdots\times [0,N_D]$ are frequencies, and $\boldsymbol{n}=(n_1, n_2, \dots, n_D)$ and $\boldsymbol{k} = (k_1, k_2, \dots, k_D)$ are $D$-dimensional vectors of indices from 0 to $\boldsymbol{N}-1 := (N_1-1, N_2-1, \dots, N_D-1).$

For the purposes of neural operator learning, we again rearrange this summation into a matrix. We store the sample points as a matrix $P = \left( \boldsymbol{p}_0, \boldsymbol{p}_1, \hdots, \boldsymbol{p}_{N-1} \right)^T \in [0,1]^{N\times D}$. Additionally, the number of frequencies, or modes, is truncated at $m$ along each spatial dimension, while higher modes are ignored. This results in the following matrix,
\begin{equation}
\label{eq:V_2d}
    \mathbf{V}_{j, k} = \sqrt{\frac{D}{N}}\left[ e^{-2 \pi i  \left( \sum\limits_{l=0}^{D-1} \left( \left\lfloor \frac{j}{m^l}\right\rfloor\mod m \right) P_{k,l} \right)} \right]_{j, k=0}^{m^D-1, N-1}.
\end{equation}
The order of the exponents is equivalent to a flattened tensor product. To further illustrate the general layout of this matrix, we present $\mathbf{V}$ for the 2D case over $m$ modes, 
\setlength{\arraycolsep}{0pt}
\begin{equation}
    \mathbf{V} = \sqrt{\frac{2}{N}}\begin{bmatrix}
    e^{-2 \pi i(0\mathbf{p_0}^T+0\mathbf{p_1}^T)} \\ e^{-2 \pi i(1\mathbf{p_0}^T+0\mathbf{p_1}^T)} \\ \vdots \\ e^{-2 \pi i((m-1)\mathbf{p_0}^T+0\mathbf{p_1}^T)} \\ e^{-2 \pi i(0\mathbf{p_0}^T+1\mathbf{p_1}^T)} \\  e^{-2 \pi i(1\mathbf{p_0}^T+1\mathbf{p_1}^T)} \\ \vdots \\ e^{-2 \pi i((m-1)\mathbf{p_0}^T+1\mathbf{p_1}^T)} \\ \vdots \\ e^{-2 \pi i(0\mathbf{p_0}^T+(m-1)\mathbf{p_1}^T)}\\ e^{-2 \pi i(1\mathbf{p_0}^T+(m-1)\mathbf{p_1}^T)} \\ \vdots \\ e^{-2 \pi i((m-1)\mathbf{p_0}^T+(m-1)\mathbf{p_1}^T)}\\
    \end{bmatrix}.
    \label{eq:flattened structured matrix}
\end{equation}
This construction is readily implemented in a single shot using tensorized methods, eliminating the need for less efficient loop constructs. Likewise, it leverages methods which have been highly optimized for 3D or 4D tensors, such as \textit{torch.bmm()} and \textit{torch.matmul()} for PyTorch.

Data from a general point cloud is stored in a vector, thus the transform relating $\mathbf{x}$ and $\mathbf{X}$ in the higher-dimensional cases is as readily calculated by the matrix-vector product (\ref{eq:transform}), as in the 1D case.

In the special case that sample points lie on a lattice, it is possible to use a more memory efficient approach which is similar to the 2D FFT. We refer the interested reader to \textbf{SM} \ref{subsec: lattice transforms} for more details.

\paragraph{Extension to Spherical Harmonics.}
The above method for constructing these matrices is not just limited to Fourier transforms and can thus be integrated in other neural operators. 
Simply using basis functions of other spectral transformations leads to a formulation that is adapted for the underlying spectral transform algorithms, such as wavelets or Laplace transforms.
We exemplify such an extension to the spherical harmonics below.

The spherical harmonics are derived as the eigenfunctions of the Laplacian on the sphere. Given the maximum degree $l_{max}$, the associated harmonics can be explicitly calculated and arranged into a matrix as,
\begin{equation}
\begin{aligned}
    \mathbf{V}_{j, k} &= \left[ C e^{i m \phi_k} \mathcal{P}^{m}_{l} (\cos \theta_k) \right]_{j,k=0}^{l_{max}^2-1, N-1}
    \\ m &= j - ( \lfloor \sqrt{j}\rfloor^2 + \lfloor \sqrt{j} \rfloor ), \quad l = \lfloor \sqrt{j} \rfloor, 
\end{aligned}
\end{equation}

for any point $k$ with polar angle $\theta_k$ and azimuth $\phi_k$, where $C$ is a normalization constant and $\mathcal{P}_l^m$ is the associated Legendre polynomial. The total number of modes is equal to  $l_{max}^2$, as each degree $l$ has $2l+1$ orders $m$, with $-m \leq l \leq m$. Once again, the transform is computed as a matrix-vector product as in (\ref{eq:transform}). 

\paragraph{Backward Transformations.}
The transformation from the spectral domain back to the physical space is immediately calculated by multiplying the data in the frequency domain by the conjugate transpose of the forward transformation matrix, 
\begin{equation}
    \mathbf{x} = \Bar{\mathbf{V}}^T \mathbf{X}.
\end{equation}
Since this approach avoids constructing a new matrix at run time, simply taking the adjoint is also computationally efficient. See Figure \ref{fig:main figure} for a visual summary of our algorithm for the realization of the \emph{discrete spectral evaluations}.

The conjugate transpose converts from the spectral to the physical domain. However, it only serves as an inverse for orthogonal forward transforms.
This is easily achieved by Euclidean Fourier transforms, when considering a sub-sampled grid where the number of sample points along each dimension is equal to the the number of Fourier modes taken along each respective dimension. In the case of spherical harmonics, however, orthogonality is preserved only in continuous cases. Nonetheless, orthogonal transformations on the sphere exist for discrete transforms under certain restrictions. \citet{driscoll1994computing} present sampling theorems for Fourier expansions and convolutions on a sphere which preserve orthogonality, as well as an efficient algorithm to compute such transforms. \citet{McEwen_2011}, likewise, present sampling theorems and fast \emph{spin} spherical harmonics algorithms for equiangular sampling points on the sphere by associating the sphere with a torus through periodic extension.


\paragraph{Computational Complexity.} The most notable feature of FFT is its computational efficiency. Calculating the Fourier coefficients of a 1D signal, sampled at $N$ points, by using the brute force DFT, costs $O(N^2)$. In contrast, the FFT algorithm computes these coefficients with $O(N \log N)$ complexity.

Hence, it is natural to wonder why one should reconsider matrix multiplication techniques in our setting. The maximum performance gain with FFT occurs when the FFT computes all the Fourier coefficients, or modes, of an underlying signal. Furthermore, peak efficiency is reached for points on a dyadic interval. While the number of modes to compute may be truncated, the interconnected nature of the self-recursive radix-2 FFT algorithm makes it difficult in practice to attain peak efficiency. We refer the reader to {\bf SM}~Figure~\ref{fig:FFT SFG} for a visual representation of the FFT algorithm. Thus, in the case of truncated modes, matrix multiplication techniques could be competitive vis a vis computational cost. As observed in \citet{barnett2019parallel}, a direct evaluation is competitive or more efficient when the number of modes is on the order of $10^1$ or fewer. 

 Moreover, for neural operators, only a small subset of nonzero modes are required to approximate the operator \citep{Li_FNO, LMHM1, lanthaler2023nonlocal}, independent of the number of points. Therefore, the computational complexity of the proposed approach cost $O(mN)$ as the  matrix structure can be fully determined in $O(mN)$ as opposed to $O(N^2)$ \citep{Gohberg, P01, GOHBERG1994411}. We also present an ablation study in {Figure~\ref{fig:ablation study}, varying the number of modes and observing the computation time for the 1D Burgers experiment, described in Section \ref{sec:Numerical Experiments}. Results for computation time as the number of modes is varied for 2D equispaced grids, 2D point clouds, and spherical geometries are presented {\bf SM} Table \ref{tab:ablation studies}. Directly evaluating the Fourier transform is clearly more efficient within the typical range of 12 to 32 modes required for by FNO \citep{Li_FNO}, and it remains more efficient even up to 64 modes. This figure suggests that direct spectral evaluations will be faster to run in practice. 

 \begin{figure}
    \centering
    \includegraphics[width=0.45\textwidth]{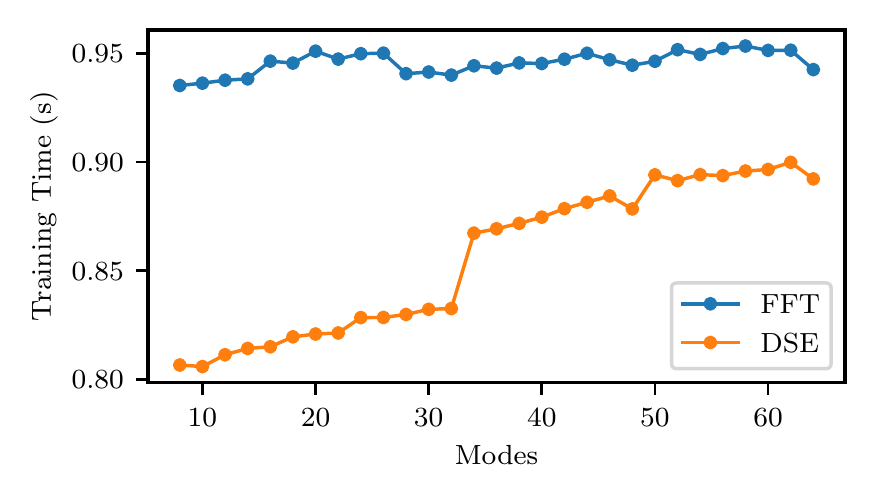}
    \vspace{-1em}
    \caption{Ablation study to compare the FNO training time using the FFT and the direct spectral evaluations (DSE) for the 1D Burgers' equation on equispaced data.}
    \label{fig:ablation study}
    \vspace{-0.5em}
\end{figure}
 
 A detailed discussion on the computational complexity of the matrix method presented here, both for Fourier transforms and Spherical harmonics is provided in {\bf SM} \ref{app:comp}. 

\section{Experimental Results}
\label{sec:Numerical Experiments}
In this section, our aim is to investigate the performance of the presented \emph{Direct Spectral Evaluations} (DSE) within various neural operator architectures on a challenging suite of diverse PDE tasks.
\paragraph{Implementation, Training Details and Baselines.} 
A key contribution of this paper is a new implementation of the presented matrix multiplications in \emph{PyTorch}, which enables us to efficiently compute Fourier transforms. Within a neural network, an efficient $O(mN)$ algorithm must also be parallelizable to handle batches, as this massively speeds up the training process. Batches of data with the same or different point distributions are easily handled by the \emph{torch.matmul()} and \emph{torch.bmm()} functions. 

In all experiments, we use a simple grid search to select the hyperparameters. We train all models until convergence. We also use the L1-loss function, which produced both a lower L1-error and L2-error than the L2-loss. The test error was measured in all experiments as the relative L1 error. 

As baselines, we use the geometric diffeomorphism (Geometric Layer), described by \citet{Li_GeoFNO} in experiments where the underlying domain has a complicated, non-equispaced geometry, when applicable. For the one dimensional experiment, we use a cubic interpolation scheme on the nonequispaced data, as well as two Nonuniform FFT approaches. For the experiments on a lattice, we take the model's performance over the original grid as a baseline. To apply the SFNO in the spherical example, we explore radial-basis function interpolation schemes using both a Gaussian kernel with variance 0.1 and a linear kernel.

To show the effectiveness as well as the generality of DSE, we implement it within several prominent neural operators, namely the FNO, UFNO \citep{UFNO}, FFNO \citep{tran2023factorized}, and SFNO \citep{SFNO} (for data on the sphere). Model sizes are chosen to be as close as possible when comparing the DSE and the baselines. All experiments are performed on the Nvidia GeForce RTX 3090 with 24GB memory.

\begin{table*}[]
    \centering
    \caption{Performance results for all experiments, comparing the DSE approach to various baselines. Directly evaluating the Fourier transform or spherical harmonics over the given domain offers clear advantages in speeding up training time, improving the testing error, or both across a variety of experiments with both equispaced grids and unusual geometries.}
    \label{tab:numerical_results}
    \resizebox{0.8\textwidth}{!}{%
    \begin{tabular}{l l c c}
        \toprule
        \textbf{Model} & \textbf{Method}  & \textbf{Training Time (per epoch)} & \textbf{L1 Test Error} \\
        \midrule
        \multicolumn{2}{l}{\textbf{1D: Burgers' Equation}}\\
        \multicolumn{2}{l}{\textit{Equispaced Distribution:}}\\
        \multirow{2}{*}{\quad FNO} & \textbf{DSE}  & \textbf{0.72}s & \textbf{0.0551}\% \\ 
        & FFT  & 0.78s & \textbf{0.0575}\% \\ 
        \multicolumn{2}{l}{\textit{Contracting-Expanding Distribution:}}\\
        \multirow{4}{*}{\quad FNO} & \textbf{DSE}  & \textbf{0.11}s & \textbf{0.184}\% \\ 
        & FFT + Cubic Interpolation  & 1.00s & {0.195}\% \\ 
        & KB-NUFFT  & 30.0s & 0.346\% \\ 
        & Toeplitz-NUFFT  & 0.85s & {0.740}\% \\ 
        \midrule
        \multicolumn{2}{l}{\textbf{2D Nonequispaced Lattice: Shear Layer}}\\
        \multirow{2}{*}{\quad FNO} & \textbf{DSE} on Noneq. Lattice       & \textbf{57}s &  \textbf{5.53}\%\\
        & FFT on Equispaced Grid        & 251s & 6.76\% \\
        \multirow{2}{*}{\quad UFNO} & \textbf{DSE} on Noneq. Lattice      & 68s &  5.61\%\\
        & FFT on Equispaced Grid       & 287s & 6.61\% \\
        \multirow{2}{*}{\quad FFNO} & \textbf{DSE} on Noneq. Lattice      & 108s & 12.4\% \\
        & FFT on Equispaced Grid      & 380s & 12.4\% \\
        \midrule
        \multicolumn{2}{l}{\textbf{2D Nonequispaced Lattice: Specific Humidity}} \\
        \multirow{2}{*}{\quad FNO} & \textbf{DSE} on Noneq. Lattice  & \textbf{1.5}s & 4.65\% \\
        & FFT on Equispaced Grid  & 15s & 5.09\% \\
        \multirow{2}{*}{\quad UFNO} & \textbf{DSE} on Noneq. Lattice  & 2.5s & \textbf{3.91}\% \\
        & FFT on Equispaced Grid  & 23s & 4.34\% \\
        \multirow{2}{*}{\quad FFNO} & \textbf{DSE} on Noneq. Lattice  & 2.3s & 4.41\% \\
        & FFT on Equispaced Grid  & 26s & 5.02\% \\
        \midrule
        \multicolumn{2}{l}{\textbf{2D Point Cloud: Flow Past Airfoil}} \\
        \multirow{2}{*}{\quad FNO} & \textbf{DSE}  & 2.8s & \textbf{0.220}\% \\
        & Geometric Layer   & 6.2s & 1.20\%\\
        \multirow{2}{*}{\quad UFNO} & \textbf{DSE}  & 2.6s & 0.380\% \\
        & Geometric Layer   & 7.1s & 0.679\% \\
        \multirow{2}{*}{\quad FFNO} & \textbf{DSE}  & \textbf{2.1}s & 0.650\% \\
        & Geometric Layer   & 6.9s & 2.10\%\\
        \bottomrule
        \multicolumn{2}{l}{\textbf{2D Point Cloud: Elasticity}} \\
        \multirow{2}{*}{\quad FNO} & \textbf{DSE}   & 0.41s & 1.96\% \\
        & Geometric Layer           & 0.71s & 2.39\%\\
        \multirow{2}{*}{\quad UFNO} & \textbf{DSE}  & 0.60s & 2.05\% \\
        & Geometric Layer          & 1.0s & 2.16\% \\
        \multirow{2}{*}{\quad FFNO} & \textbf{DSE}  & \textbf{0.28}s & \textbf{1.73}\% \\%
        & Geometric Layer          & 0.44s & 2.20\%\\
        \bottomrule
        \multicolumn{2}{l}{\textbf{Random Spherical Point Cloud: Shallow Water Equations}} \\
        \multirow{3}{*}{\quad SFNO} & \textbf{DSE}& \textbf{15}s & \textbf{3.88}\% \\
        & Gaussian Interpolation  & 86s & 7.29\%\\
        & Linear Interpolation  & 71s & 12.7\%\\
        \multirow{3}{*}{\quad FNO} & \textbf{DSE}  & 16s & 5.39\% \\
        & Gaussian Interpolation     & 92s & 8.41\%\\
        & Linear Interpolation       & 83s & 15.2\%\\
        \bottomrule
    \end{tabular}
    }
\end{table*}
\paragraph{Benchmark 1: Burgers' Equation.}
The one-dimensional viscous Burgers' equation is a widely considered model problem for fluid flow given by
\begin{equation}
    \begin{split}
        \partial_t u(x,t) + \partial_x \left(\frac{1}{2}u^2(x,t)\right) & = \nu \partial_{xx} u(x,t), \\
        u(x,0) & = u_0(x), \\
    \end{split}
\end{equation}
where $x \in (0,1), t \in (0,1]$, $u$ denotes the fluid velocity and $\nu$ the viscosity. We follow \citep{Li_FNO} in fixing $\nu = 0.1$ and considering the operator that maps the initial data $u_0$ to the solution $u(\cdot,T)$ at final time $T=1$. The training and test data, presented in \citep{Li_FNO} for this problem, is used. We start by comparing the standard version of FNO (with FFT) to FNO with the DSE for data sampled on an equispaced grid. The difference between the resulting test errors is negligible, because the underlying algorithm is the same in this case modulo small numerical errors, as reported in Table \ref{tab:numerical_results}. Moreover, the training times per epoch were also comparable. In contrast, for data sampled from points drawn from a contracting-expanding distribution, illustrated in {\bf SM}~Figure \ref{fig:burgers distributions}, the proposed method was $5\%$ more accurate on average when compared to FNO with a cubic interpolation, interpolating data from the contracting-expanding distribution to an equispaced grid. However, the training time with the DSE was notably improved, by almost a factor of $4$, when compared to the interpolation baseline. Note that the cubic interpolation was computed beforehand and its cost not taken into account in reporting the training time. In addition, we draw comparisons with relevant Nonuniform FFT (NUFFT) algorithms with PyTorch implementations, namely the Kaiser-Bessel and Toeplitz NUFFT \citep{muckley2020}. The performance of these approaches is relatively poor, as they fundamentally rely on an interpolation+FFT scheme.

\paragraph{Benchmark 2: Shear Layer.} We follow a recent work on convolutional neural operators~\citep{CNO} in considering the incompressible Navier-Stokes equations
\begin{equation}
 \label{eq:IncompressibleEuler}
    \frac{\partial \mathbf{u}}{\partial t} + \mathbf{u}\cdot{\nabla \mathbf{u}} + \nabla p = \nu \Delta \mathbf{u}, \quad  \nabla \cdot{ \mathbf{u}} = 0.
\end{equation}
Here, $\mathbf{u} \in \mathcal{R}^2$ is the fluid velocity and $p$ is the pressure. The underlying domain is the unit square with periodic boundary conditions and the viscosity $\nu = 4\times 10^{-4}$, only applied to high-enough Fourier modes (those with amplitude $\geq 12$) to model fluid flow at \emph{very high Reynolds-number}. The solution operator maps the initial conditions $\mathbf{u}(t=0)$ to the solution at final time $T=1$. We consider initial conditions representing the well-known \emph{thin shear layer} problem \citep{BCG1,LMP1} (see \citep{CNO} for details), where the shear layer evolves via vortex shedding to a complex distribution of vortices (see {\bf SM} Figure~\ref{fig:shear layer predictions} for an example of the flow). The training and test samples are generated, with a spectral viscosity method \citep{LMP1} of a fine resolution of $1024^2$ points, from an initial sinusoidal perturbation of the shear layer \citep{LMP1}, with layer thickness of $0.1$ and $10$ perturbation modes, the amplitude of  each sampled uniformly from $[-1,1]$ as suggested in \citep{CNO}. As seen from {\bf SM} Figure~\ref{fig:shear layer predictions}, the flow shows interesting behavior with sharp gradients in two mixing regions, which are in the vicinity of the initial interfaces. However, the flow is nearly constant further away from this mixing region. Hence, we will consider input functions being sampled on a lattice shown in {\bf SM} Figure~\ref{fig:shear flow distributions} which is adapted to resolve regions with large gradients of the flow. On the other hand, the FFT based FNO, UFNO and FFNO baselines are tested on the equispaced point distribution. From Table \ref{tab:numerical_results}, we observe that the proposed method is marginally more accurate while consistently being $4$ times faster per training epoch across all models, demonstrating a significant computational advantage on this benchmark. 
\paragraph{Benchmark 3: Surface-Level Specific Humidity.} Next, we focus on a \emph{real world} data set and learning task where the objective is to predict the surface-level specific humidity over South America at a later time ($6$ hours into the future), given inputs such as wind speeds, precipitation, evaporation, and heat exchange at a given time. The exact list of inputs is given in {\bf SM}~Table~\ref{tab:humidity parameters}. The physics of this problem are intriguingly complex, necessitating a \emph{data-driven approach} to learn the underlying operator. To this end, we use the (MERRA-2) satellite data to forecast the surface-level specific humidity~\citep{NASAEarthData}. Moreover, we are interested in a more accurate regional prediction, namely over the Amazon rainforest. Hence, for models with the DSE, we will sample data on points on a lattice that is more dense over this rainforest, while being sparse (with smooth transitions) over the rest of the globe, see {\bf SM}~Figure~\ref{fig:humidity distributions} for visualization of this lattice. Test error is calculated over the region, shown in {\bf SM} Figure~\ref{fig:humidity predictions}. The results, presented in Table~\ref{tab:numerical_results}, show that the localization capabilities of the proposed method not only offer more accurate results, but DSE based neural operators are also one order of magnitude faster to train than the baselines. The greater accuracy is also clearly observed in {\bf SM} Figure~\ref{fig:humidity predictions}, where we observe that the FNO-DSE is able to capture elements such as the formation of vortices visible in the lower right hand corner and the mixing of airstreams over the Pacific Ocean. 
\paragraph{Benchmark 4: Flow Past Airfoil.}
We also investigate transonic flow over an airfoil, as governed by the compressible Euler equations,
with the problem setup  considered in \citep{Li_GeoFNO}. The underlying operator maps the airfoil shape to the pressure field. In this case, the underlying distribution of sample points changes between each input (airfoil shape). Directly formulating the Fourier transform within neural operators, one can readily handle this situation. As baselines, we augment FNO, UFNO and FFNO with the geometric layer of Geo-FNO as proposed in \citep{Li_GeoFNO}. To have a fair comparison with DSE based models, we allow Geometric layer based models to learn the underlying diffeomorphism online. The test errors, presented in Table \ref{tab:numerical_results} show that DSE-based models are both much more accurate and much faster to train than the geometric layer based baselines.  

\paragraph{Benchmark 5: Elasticity-P.}
Again, we follow \citep{Li_GeoFNO}, to investigate the performance of this method on hyper-elastic materials. We take the data, exactly as outlined in their work, predicting the stress as output. Again, we train all models (FNO, UFNO, FFNO) on the point cloud data with DSE and with the geometric layer (as proposed in \citep{Li_GeoFNO}) as a baseline. For each underlying model, the DSE is more accurate than the geometric layer while being significantly faster to train. 

\paragraph{Benchmark 6: Spherical Shallow Water Equations.}

We follow the recent work on spherical neural operators \citep{SFNO} by considering the shallow water equations on a rotating sphere. In this experiment, training data is generated on the fly such that new data is used for training and evaluation in every epoch, as described in \citep{SFNO}. For each sample, we draw points from a random uniform distribution over the sphere to create a point cloud, as opposed to a grid, see {\bf SM} Figure \ref{fig:spherical distributions} for an illustration of this point cloud. These may correspond to sensors over the globe in real-world applications. The training data for both models each have approximately 5000 points. As baselines, we considered SFNO but with data interpolated back to a regular grid on the sphere using either linear interpolation or radial basis function interpolation with a Gaussian kernel. We also compare the FNO on the interpolated data with the DSE approach on the point cloud data. The results, presented in Table \ref{tab:numerical_results}, clearly show that SFNO using DSE readily outperforms baselines on the interpolated data in terms of both accuracy as well as training speed. This difference is very pronounced for the baselines based on interpolation, which are both expensive as well as inaccurate. As expected the FNO-DSE model performs significantly worse than the SFNO-DSE model which takes into account the underlying spherical geometry. Additionally, we provide the baseline comparison of the SFNO and FNO on the original (not from interpolation) grid in \textbf{SM} Table \ref{tab:sfno original grid results} as a reference to their relative performance. Likewise, we test a trained SFNO model on samples with varying numbers of collocation points. These results, presented in Table \ref{tab:sfno smm varying grids results} show that this method is capable of retaining performance, even when trained on grids at different resolutions. 

Summarizing, the results of the numerical experiments clearly demonstrate that DSE is able to handle data sampled on arbitrary point distributions and to accurately approximate the underlying operators. It readily outperforms baselines, based either on interpolation or geometric transformations, both on accuracy as well as on training speed. This is particularly evident on problems where the data is sampled on point clouds and where other operator learning methods such as U-Nets or convolutional neural operators cannot be readily applied.

\section{Discussion}
\label{sec:discussion}

\textbf{Summary.}
Widely used neural operators such as FNO and its variants are limited to input/output data sampled on equispaced grids as the underlying spectral transformations can only be efficiently evaluated on this data structure. To expand the range of such neural operators, we propose replacing the standard computations of these spectral transformations within neural operators with a simple, general method that leverages the low number of (truncated) modes to efficiently compute the spectral transform even on arbitrary sampling point distributions (point clouds). By design, our method is able to process data on arbitrary point distributions with low computational cost. A novel PyTorch implementation is also presented to realize the proposed approach that we term as the Direct Spectral Evaluation (DSE). 

DSE is flexible enough to be embedded within any neural operator that uses non-local spectral transformations. We test with a variety of neural operators such as FNO, UFNO, FFNO, and recently SFNO for data on a sphere. Moreover, DSE is general enough to handle point distributions ranging from equispaced grids through lattices to randomly distributed point clouds. We investigate the efficiency of the proposed DSE by testing it in conjunction with a variety of available neural operators on a suite of benchmarks that correspond to a variety of PDEs as well as sampling point distributions. In all the considered benchmarks, DSE outperformed the baselines (based either on interpolation or learnable geometric differomorphisms as proposed in \citep{Li_GeoFNO}) both in terms of accuracy as well as training speed, even showing order of magnitude speedups in some cases. Additionally, we present in \textbf{SM} Table \ref{tab:sfno smm varying grids results}, the generalization capabilities of the DSE approach to be applied to point clouds not encountered during training time, even when the number of points varies greatly. Thus, we present a novel yet simple method that can be used readily within neural operators to handle input/output data sampled on arbitrary point distributions and learn the underlying operators accurately and efficiently.

\textbf{Related Work.}
Nonequispaced FFT (NUFFT) algorithms have already been developed \citep{dutt1993, beylkin1995, dutt1995, Liu_NUFFT, Kircheis_Study} as well as libraries for efficient GPU implementations \citep{shih2021cufinufft} and PyTorch implementations \citep{muckley2020}. 
The techniques frequently used in these NUFFTs or approximated inverse NUFFT include interpolation, windowing, and/or sampling along with FFT to achieve $\mathcal{O}(n \log n)$ complexity algorithms \citep{selva2018, heinig1984, gelb2014, kunis2007, ruizantolin2018,averbuch2016, kircheis2019, Kircheis_Inverse, Kunis_NFFT, perera2022}.
Essentially, the NUFFT involves interpolating to the equispaced grid followed by the FFT. However, interpolation based-approaches has already been shown to be sub-optimal for the case of neural operator learning on arbitrary \citep{Li_GeoFNO}. Furthermore, interpolation within the operator structure increases the computation time, making it more efficient to interpolate data to a grid before training or testing and simply apply the standard FNO. The NUDFT is rarely used, as many applications of Fourier transforms require all Fourier coefficients, resulting in $O(n^2)$ cost~\citep{Bagchi_NUDFT}. However, this is not the case for the neural operators that we consider, and thus the proposed DSE avoids the rapidly growing computational costs associated with NUDFT. In some implementations, the Geo-FNO also uses a nonequispaced Fourier transform in conjunction with the diffeomorphism. As opposed to constructing a standard matrix, \citep{Li_GeoFNO} construct an $N+1$ tensor, $N$ being the number of spatial dimensions. This geometric-Fourier layer then transforms the problem to a set of points on a regular grid, where the FFT is used.  While the nonequispaced transformation does have some similarities to that proposed here, the proposed method differs fundamentally in its use, leading to notable differences in the results. The construction and multiplication we offer in our implementation are fast and efficient, allowing us to use the nonequispaced transformation within each Fourier layer, as opposed to transforming the data to an equispaced grid through some slower transformation process and then apply the FFT. Likewise, this method allows us to avoid the need for any sort of diffeomorphism. In practice, we found the diffeomorphism layer difficult to train in tandem with the Fourier layer of the FNO. This process often requires careful tuning of the associated diffeomorphism loss, as well as the need to freeze this layer after some point of time and to retrain the FNO with the diffeomorphism model fixed. In contrast, the FNO-DSE converges just as reliably as the basic FNO and we see the same behavior with variants of FNO such as UFNO and FFNO. 

\textbf{Limitations and Future Work.}
The elements of the matrix are directly related to the positions of the data points for a given problem. Thus, if all training samples have different point clouds, a different matrix must be constructed for each sample. Constructing the matrices at run time, \textit{i.e.}, during training, hinders performance; however, this is mitigated by the tensorized matrix construction methods available in the implementation. As outlined by the results, the run-time matrix construction is still able to outperform other techniques for handling point cloud data. Finally, we would like to highlight the potential of the DSE for dealing with very general spectral transformations for data sampled on arbitrary point distributions. We have considered both Fourier transforms and Spherical harmonics in this paper but other relevant spectral transforms such as Wavelets or Laplace transforms will also be considered in future work. 

\newpage

\section*{Impact Statement}
This paper presents work whose goal is to advance the field of 
Machine Learning, specifically for applications in scientific computing. There are many potential societal consequences 
of our work, none of which we feel must be specifically highlighted here.

\section*{Acknowledgement}
Sirani M. Perera's work was partially supported by the NSF award number 2229473.

\bibliography{SMM_Paper}
\bibliographystyle{icml2024}

\newpage
\appendix
\begin{center}
{\bf Supplementary Material for:} \\ 
Fourier-Based Neural Operators on Arbitrary Domains\\
\end{center}

\section{Technical Details} 

\subsection{The FFT Signal Flow Graph.} The signal flow graph provides a graphical representation of the FFT algorithm. Radix-2, recursive algorithms such as the FFT are highly dependent on the positions of the data. Violating this principle makes using similar approaches difficult, as the processes can not be easily "untangled" in some sense.
\begin{figure}[h]
    \begin{center}

\begin{tikzpicture}

    \matrix[row sep=5mm, column sep=10mm]
    {
		\node[dspnodeopen, dsp/label = left] 	(m{0,0}) {$x(0)$};&
		\node[dspnodeopen]                  (m{0,1}) {};          &
		\node[dspnodeopen]                 	(m{0,2}) {};          &
		\node[dspnodeopen]                  (m{0,3}) {}; 	&
		\node[dspnodeopen]                 	(m{0,4}) {};          &
		\node[dspnodeopen, dsp/label = right]  (m{0,5}) {$y(0)$};   &\\
        \node[dspnodeopen, dsp/label = left] 	  (m{1,0}) {$x(1)$};&
		\node[dspnodeopen]                        (m{1,1}) {};          &
		\node[dspnodeopen]                 	      (m{1,2}) {};          &
		\node[dspnodeopen]                        (m{1,3}) {}; 	&
		\node[dspnodeopen]                 	      (m{1,4}) {};          &
		\node[dspnodeopen, dsp/label = right]     (m{1,5}) {$y(2)$};   &\\
		\node[dspnodeopen, dsp/label = left] 	  (m{2,0}) {$x(2)$};&
		\node[dspnodeopen]                        (m{2,1}) {};          &
		\node[dspnodeopen]                 	      (m{2,2}) {};          &
		\node[dspnodeopen]                        (m{2,3}) {}; 	&
		\node[dspnodeopen]                 	      (m{2,4}) {};          &
		\node[dspnodeopen, dsp/label = right]     (m{2,5}) {$y(4)$};   &\\
		\node[dspnodeopen, dsp/label = left] 	  (m{3,0}) {$x(3)$};&
		\node[dspnodeopen]                        (m{3,1}) {};          &
		\node[dspnodeopen]                 	      (m{3,2}) {};          &
		\node[dspnodeopen]                        (m{3,3}) {}; 	&
		\node[dspnodeopen]                 	      (m{3,4}) {};          &
		\node[dspnodeopen, dsp/label = right]     (m{3,5}) {$y(6)$};   &\\
		\node[dspnodeopen, dsp/label = left] 	  (m{4,0}) {$x(4)$};&
		\node[dspnodeopen]                        (m{4,1}) {};          &
		\node[dspnodeopen]                 	      (m{4,2}) {};          &
		\node[dspnodeopen]                        (m{4,3}) {}; 	&
		\node[dspnodeopen]                 	      (m{4,4}) {};  &        
		\node[dspnodeopen, dsp/label = right]     (m{4,5}) {$y(1)$};   &\\
		\node[dspnodeopen, dsp/label = left] 	  (m{5,0}) {$x(5)$};&
		\node[dspnodeopen]                        (m{5,1}) {};          &
		\node[dspnodeopen]                 	      (m{5,2}) {};          &
		\node[dspnodeopen]                        (m{5,3}) {}; 	&
		\node[dspnodeopen]                 	      (m{5,4}) {};          &
		\node[dspnodeopen, dsp/label = right]     (m{5,5}) {$y(3)$};   &\\
		\node[dspnodeopen, dsp/label = left] 	  (m{6,0}) {$x(6)$};&
		\node[dspnodeopen]                        (m{6,1}) {};          &
		\node[dspnodeopen]                 	      (m{6,2}) {};          &
		\node[dspnodeopen]                        (m{6,3}) {}; 	&
		\node[dspnodeopen]                 	      (m{6,4}) {};          &
		\node[dspnodeopen, dsp/label = right]     (m{6,5}) {$y(5)$};   &\\
		\node[dspnodeopen, dsp/label = left] 	  (m{7,0}) {$x(7)$};&
		\node[dspnodeopen]                        (m{7,1}) {};          &
		\node[dspnodeopen]                 	      (m{7,2}) {};          &
		\node[dspnodeopen]                        (m{7,3}) {}; 	&
		\node[dspnodeopen]                 	      (m{7,4}) {};          &
		\node[dspnodeopen, dsp/label = right]     (m{7,5}) {$y(7)$};   &\\
    };
    
\draw[dspflow] (m{0,0}) -- (m{0,1});
\draw[dspflow] (m{1,0}) -- (m{1,1});
\draw[dspflow] (m{2,0}) -- (m{2,1});
\draw[dspflow] (m{3,0}) -- (m{3,1});
\draw[dspflow,dashed] (m{4,0}) -- (m{4,1});
\draw[dspflow,dashed] (m{5,0}) -- (m{5,1});
\draw[dspflow,dashed] (m{6,0}) -- (m{6,1});
\draw[dspflow,dashed] (m{7,0}) -- (m{7,1});

\draw[dspconn] (m{0,0}) -- (m{4,1});
\draw[dspconn] (m{1,0}) -- (m{5,1});
\draw[dspconn] (m{2,0}) -- (m{6,1});
\draw[dspconn] (m{3,0}) -- (m{7,1});
\draw[dspconn] (m{4,0}) -- (m{0,1});
\draw[dspconn] (m{5,0}) -- (m{1,1});
\draw[dspconn] (m{6,0}) -- (m{2,1});
\draw[dspconn] (m{7,0}) -- (m{3,1});
\draw[dspflow] (m{0,1}) -- (m{0,2});
\draw[dspflow] (m{1,1}) -- (m{1,2});
\draw[dspflow] (m{2,1}) -- (m{2,2});
\draw[dspflow] (m{3,1}) -- (m{3,2});
\draw[dspflow] (m{4,1}) -- (m{4,2});
\draw[dspflow] (m{5,1}) -- node[above] {\color{red}$e^{\frac{-i \pi}{4} }$}(m{5,2});
\draw[dspflow] (m{6,1}) -- node[above] {\color{red}$-i$}(m{6,2});
\draw[dspflow] (m{7,1}) -- node[above] {\color{red}$e^{\frac{-3 i \pi}{4} }$}(m{7,2});

\draw[dspflow]          (m{0,2}) -- (m{0,3});
\draw[dspflow]          (m{1,2}) -- (m{1,3});
\draw[dspflow,dashed]   (m{2,2}) -- (m{2,3});
\draw[dspflow,dashed]   (m{3,2}) -- (m{3,3});
\draw[dspflow]          (m{4,2}) -- (m{4,3});
\draw[dspflow]          (m{5,2}) -- (m{5,3});
\draw[dspflow,dashed]   (m{6,2}) -- (m{6,3});
\draw[dspflow,dashed]   (m{7,2}) -- (m{7,3});

\draw[dspconn] (m{0,2}) -- (m{2,3});
\draw[dspconn] (m{1,2}) -- (m{3,3});
\draw[dspconn] (m{2,2}) -- (m{0,3});
\draw[dspconn] (m{3,2}) -- (m{1,3});
\draw[dspconn] (m{4,2}) -- (m{6,3});
\draw[dspconn] (m{5,2}) -- (m{7,3});
\draw[dspconn] (m{6,2}) -- (m{4,3});
\draw[dspconn] (m{7,2}) -- (m{5,3});

\draw[dspflow] (m{0,3}) -- (m{0,4});
\draw[dspflow] (m{1,3}) -- (m{1,4});
\draw[dspflow] (m{2,3}) -- (m{2,4});
\draw[dspflow] (m{3,3}) -- node[above] {\color{red}$-i$}(m{3,4});
\draw[dspflow] (m{4,3}) -- (m{4,4});
\draw[dspflow] (m{5,3}) -- (m{5,4});
\draw[dspflow] (m{6,3}) -- (m{6,4});
\draw[dspflow] (m{7,3}) -- node[above] {\color{red}$-i$}(m{7,4});

\draw[dspflow]          (m{0,4}) -- (m{0,5});
\draw[dspflow,dashed]   (m{1,4}) -- (m{1,5});
\draw[dspflow]          (m{2,4}) -- (m{2,5});
\draw[dspflow,dashed]   (m{3,4}) -- (m{3,5});
\draw[dspflow]          (m{4,4}) -- (m{4,5});
\draw[dspflow,dashed]   (m{5,4}) -- (m{5,5});
\draw[dspflow]          (m{6,4}) -- (m{6,5});
\draw[dspflow,dashed]   (m{7,4}) -- (m{7,5});

\draw[dspconn] (m{0,4}) -- (m{1,5});
\draw[dspconn] (m{1,4}) -- (m{0,5});
\draw[dspconn] (m{2,4}) -- (m{3,5});
\draw[dspconn] (m{3,4}) -- (m{2,5});
\draw[dspconn] (m{4,4}) -- (m{5,5});
\draw[dspconn] (m{5,4}) -- (m{4,5});
\draw[dspconn] (m{6,4}) -- (m{7,5});
\draw[dspconn] (m{7,4}) -- (m{6,5});

\end{tikzpicture}

    \end{center}
    \caption{The 8-point fast Fourier transform signal flow graph. $x$ and $y$ represent the signal in the physical and Fourier domain, respectively. Dashed lines represent a multiplication by -1, red elements denote a multiplication by that factor, and converging arrows represent a sum.}
    \label{fig:FFT SFG}
\end{figure}

\subsection{Transformations on the 2D Lattice}
\label{subsec: lattice transforms}
The 2D FFT is calculated as two 1D FFTs along each axis, as 
\begin{equation}
    X = \mathcal{F}x \mathcal{F}^T,
\end{equation} where $\mathcal{F}$ is the FFT, transforming an array $x \in \mathbb{C}^{N_0\times N_1}$ to an array $X \in \mathbb{C}^{N_0\times N_1}$. $N_0$ and $N_1$ correspond to the number of points along the first and second spatial axes, respectively.

We may mirror this transformation on any 2D lattice, i.e, the tensor product of 1D point distributions along each axis (see {\bf SM} Figure~\ref{fig:Point Distributions}). To do this, we construct two matrices, $\mathbf{V}_0 \in \mathbb{C}^{m \times N_0}$ and $\mathbf{V}_1 \in \mathbb{C}^{N_1 \times m}$, corresponding to the sample points of data points along the first and second axis, respectively. 
\begin{equation}
    X= \mathbf{V}_1 x \mathbf{V}_2^T.
    \label{eq:V_2d_lattice}
\end{equation}
The operational complexity of this approach is equivalent to that of Section 2; however, the matrices contain only $m \times (N_0 + N_1)$ values, while those in the general case contain $m^2 \times (N_0 \times N_1)$ values. Therefore, this approach may be more efficient depending on the distribution of sample points.

\subsection{On Computational Complexity of the Direct Evaluation of Fourier Transforms.}
\label{app:comp}
The matrix-vector product, as described in Section 2 of the main text, is broken down into a  vector with $N$ components via inner products of vectors, requiring $2N$ real multiplications and $2(N-1)$ real additions having a real-valued input, a total of $4N^2 - 2n$ flops. In the implementation, the number of rows is reduced from $N$ to a constant $m$ s.t. $m<<N$, as determined by the number of modes chosen for a given problem. Thus, for the 1D case, the total number of flops is $4Nm - 4N$, therefore only growing linearly with the problem size. For the $D$-dimensional nonequispaced data, we construct a matrix with size $N$ columns but the number of rows $m_{total} = (m_1)(m_2)\hdots(m_D)$, where $m_j$ is the number of modes taken along the $j^{th}$ spatial dimension. Thus, the complexity of the transform using the direct evaluation is reduced from $\mathcal{O}(N^2)$ to $\mathcal{O}(m_{total}N)$. Finally, we note here that the existing non-uniform FFT has the complexity of order $\mathcal{O}(R N \: log(N))$, where $R$ is based on diagonally scaled FFT. Thus, applying the existing non-uniform FFT on FNO won't reduce the complexity of the direct evaluation on the neural operator with the complexity reduction of order $\mathcal{O}(m_{total}N)$.

For the extension to the spherical harmonics, we pre-compute the Legendre polynomials over $N$ points as an evaluation problem with  $\mathcal{O}(N)$ complexity. Once these are pre-computed, we call these polynomials to calculate the spherical harmonics with complexity $\mathcal{O}{(l^2)}$, e.g. the order $l$ is calculated using an odd number of harmonics from 1 to the order $l$. Thus, the associated spherical harmonics can be computed using $\mathcal{O}{(N \:l^2)}$ complexity, where $l << N$. Finally, arranging harmonics into the truncated matrix form followed by the matrix-vector products into vector forms (as described above) reduces the complexity to $\mathcal{O}(N l^2)$ as opposed to $\mathcal{O}{(N^2 \:l^3)}$.

The question may arise; \textit{Can the number of modes really be said to be constant with respect to the number of points, as more points may be used for more complex problems, necessitating more modes?} A priori, it is not possible to say how many modes must be selected for a problem of given complexity or resolution - these must be determined through model selection, varying the number of modes, and choosing the model that minimizes the error over the validation set. This model selection process has revealed that there is often a point where increasing the number of modes results in worse performance, i.e. increasing the number of modes will not automatically result in increased performance. With regard to the sharp features and large gradients that may arise in complex sets of PDEs or high-resolution data, the lifting layer and the skip connections of the model architecture are much more capable of resolving such features than the Fourier layers. The Fourier layers serve to propagate information throughout the domain. Thus, we maintain the position that the number of modes is fixed with respect to the number of points.

\section{Experimental Details} 
\subsection{Point Distributions}
We investigate both a uniform and a \textit{contracting-expanding} distribution to help lay the foundation for the proposed method. These distributions are visualized in Figure \ref{fig:burgers distributions}. 

For the two dimensional experiments, we investigate lattices as well as random or structured data on a point cloud, simplified illustrations of which are provided in Figure \ref{fig:Point Distributions}. A lattice with a nonuniform distribution along one axis (\emph{Shear Layer}) and a lattice with a nonuniform distribution along both axes (\emph{Surface-Level Specific Humidity}) are investigated and visualized in Figures \ref{fig:shear flow distributions} and \ref{fig:humidity distributions}. Point cloud data with varying geometries are investigated in the \emph{Airfoil} and \emph{Elasticity} experiments. We provide visualizations of several airfoil point clouds in Figure \ref{fig:airfoil distributions}. 

Finally, we investigate the performance of spherical harmonics models on a random distribution over the sphere. The original point cloud, the random points, and the grid generated by interpolating from the random points are shown in Figure \ref{fig:spherical distributions}.

\begin{figure*}
    \centering
    \includegraphics[width=0.8\textwidth]{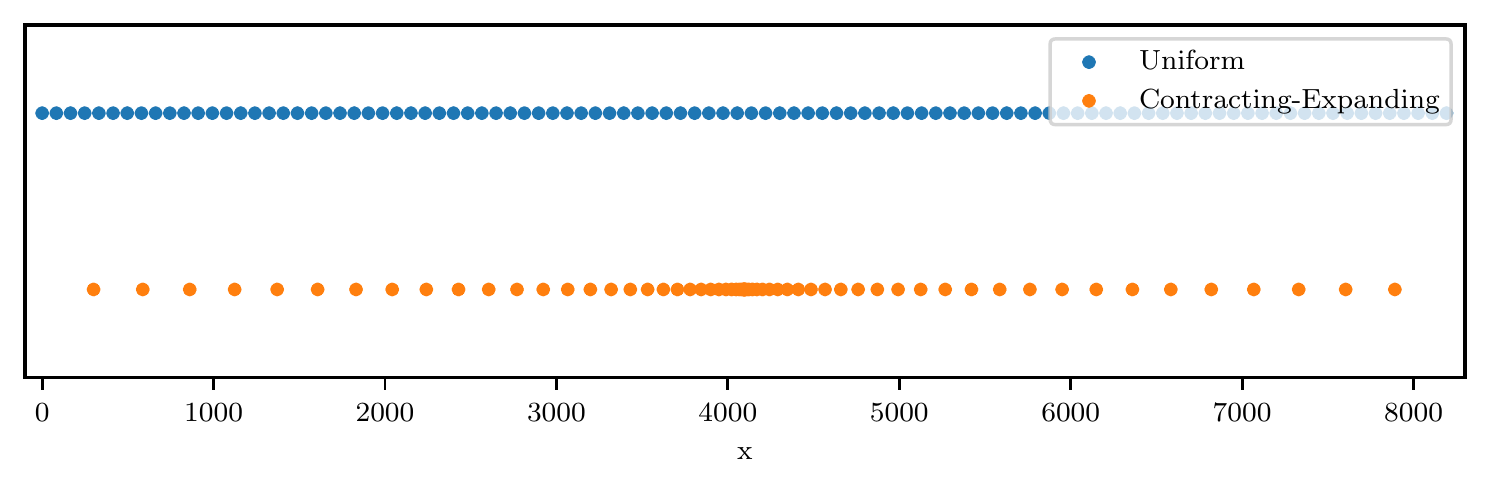}
    \caption{Point distributions used in the Burgers' equation experiments. Data is selected from the uniform distribution to construct the contracting-expanding distribution and random distribution. The space between points in the contracting-expanding distribution grows from a point in both directions according to a geometric distribution.}
    \label{fig:burgers distributions}
\end{figure*}


\begin{figure*}
    \centering
    \includegraphics[width=0.55\textwidth]{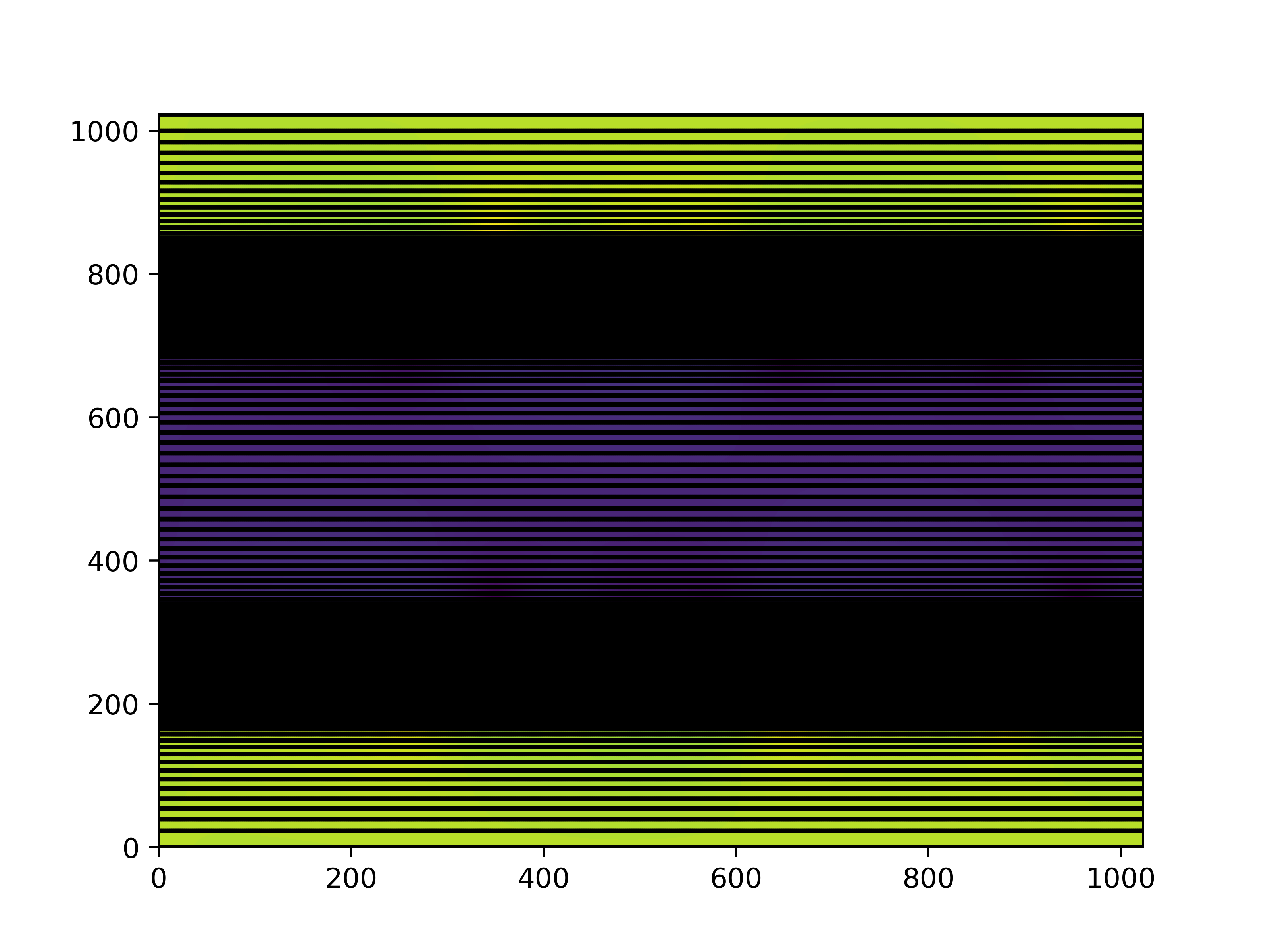}
    \caption{Nonequispaced lattice for the shear layer problem. Sampling is dense close the interface region, smoothly becoming sparse further from this region.}
    \label{fig:shear flow distributions}
\end{figure*}

\begin{figure*}
\centering
\begin{subfigure}{0.45\textwidth}
    \centering
    \includegraphics[width=\linewidth]{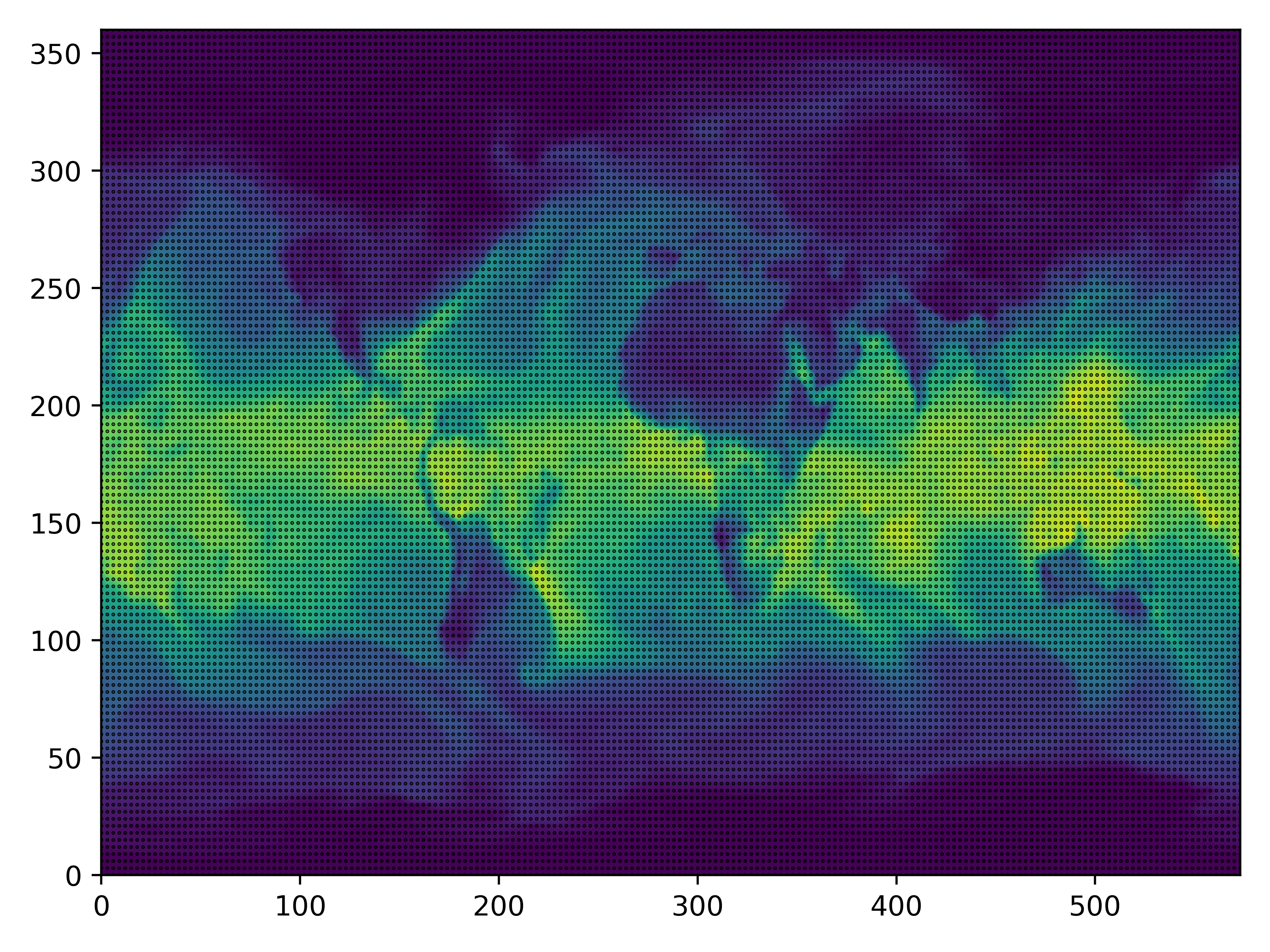}
    \caption{FNO point distribution. The points displayed in this image have been subsampled from the original distribution to maintain clarity in the figure.}
\end{subfigure}
\hfill
\begin{subfigure}{0.45\textwidth}
    \centering
    \includegraphics[width=\linewidth]{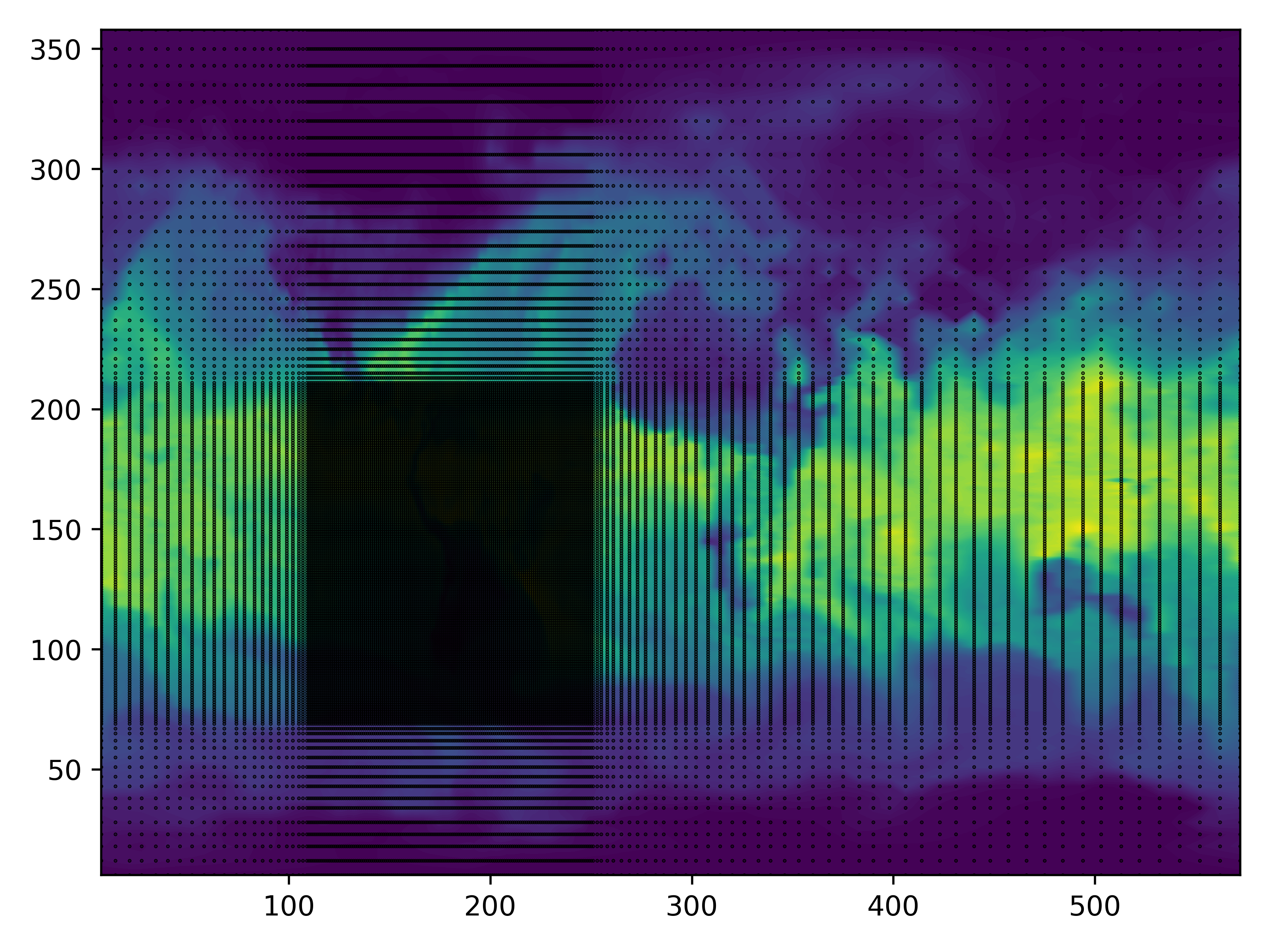}
    \caption{Nonequispaced lattice point distribution. A densely sampled region is located over South America and the lattice becomes more sparse further from this region.}
\end{subfigure}
\caption{Distributions used within the surface-level specific humidity experiment.}
\label{fig:humidity distributions}
\end{figure*}

\begin{figure*}[!h]
\centering
\begin{subfigure}{0.3\textwidth}
    \centering
    \includegraphics[width=\linewidth]{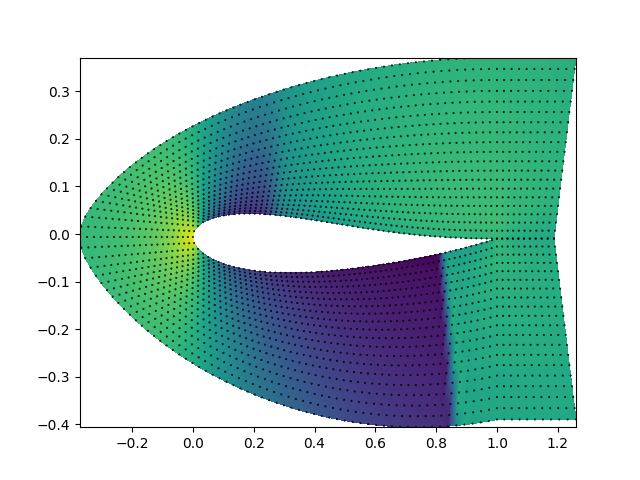}
    \caption{}
\end{subfigure}
\hfill
\begin{subfigure}{0.3\textwidth}
    \centering
    \includegraphics[width=\linewidth]{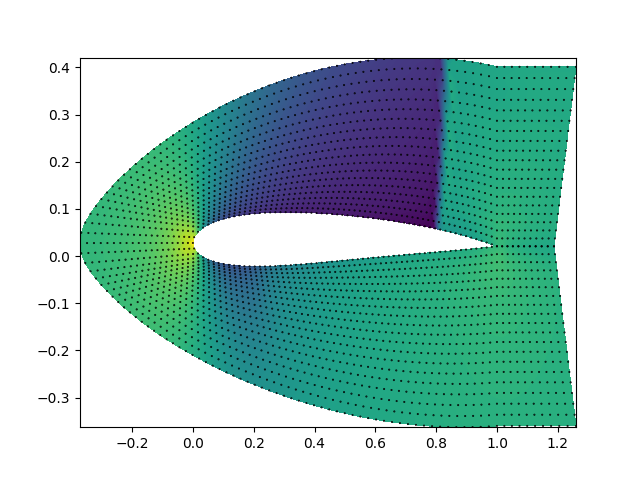}
    \caption{}
\end{subfigure}
\hfill
\begin{subfigure}{0.3\textwidth}
    \centering
    \includegraphics[width=\linewidth]{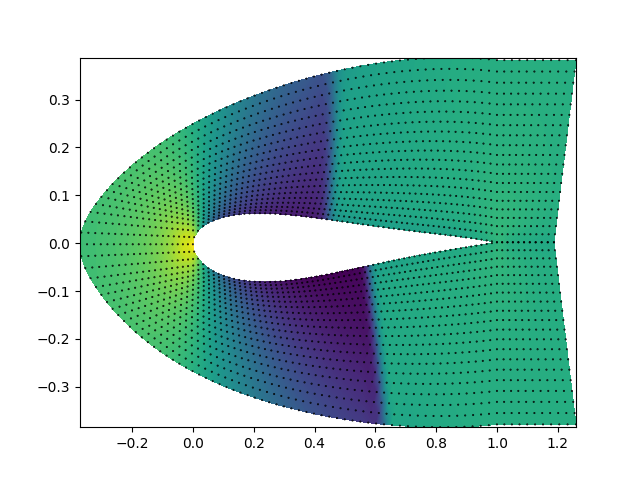}
    \caption{}
\end{subfigure}
\caption{Several point distributions around airfoils and their associated pressure distributions.}
\label{fig:airfoil distributions}
\end{figure*}

\begin{figure*}
    \centering
    \includegraphics[width=1.0\textwidth]{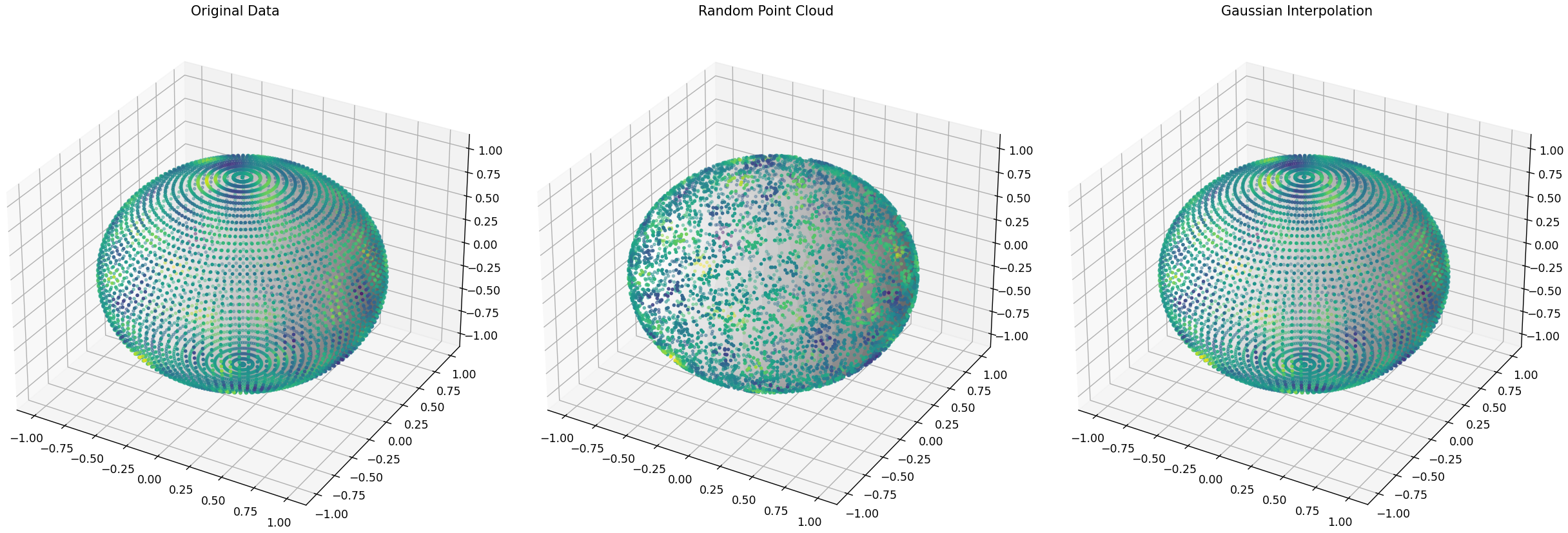}
    \caption{The original distribution of points (left) is sampled from at locations chosen randomly along the surface of a sphere (middle). To compare with baselines, which require a uniform distribution, we use a radial-basis function interpolation scheme with a Gaussian kernel and a variance of 0.1 to interpolate from the random points back up to a grid (right) as well as a linear interpolation schemer.}
    \label{fig:spherical distributions}
\end{figure*}

\newpage
\subsection{Additional Training Details and Results.}

We use 16 different parameters to make predictions for the \emph{Surface-Level Specific Humidity} experiments. These are listed in this subsection in Table \ref{tab:humidity parameters}. We also show the predictions of several models for several experiments in Figure \ref{fig:predictions}. Table \ref{tab:burgers initializations} provides information on the distributions of the minimum median relative L1 test error given different initial seeds. We also provide histograms showing the error for all test samples in Figures \ref{fig:burgers histograms} and \ref{fig:2d histograms}. We present additional results for the training times as a function of the number of modes for the Burgers, 2D Specific Humidity, 2D Airfoil, and Spherical Shallow Water experiments in Table \ref{tab:ablation studies}. Finally, we include Table \ref{tab:models sizes} to show the sizes of the various models.

\begin{table*}[!h]
    \centering
    \caption{Inputs for the surface level specific humidity predictions.}
    \label{tab:humidity parameters}
    \begin{tabular}{l l c}
    \toprule
    \textbf{Acronym in MERRA-2} & \textbf{Input} & \textbf{Units}\\
    \midrule
    CDH & heat exchange coefficient & $\frac{kg}{m^2s}$ \\
    CDQ & moisture exchange coefficient  & $\frac{kg}{m^2s}$ \\
    EFLUX & total latent energy flux  & $\frac{W}{m^2}$ \\
    EVAP & evaporation from turbulence  & $\frac{kg}{m^2s}$ \\
    FRCAN & areal fraction of anvil showers  & 1\\
    FRCCN & areal fraction of convective showe & 1\\
    FRCLS & areal fraction of large scale show & 1\\
    HLML & surface level height  & $m$\\
    QLML & surface level specific humidity  & 1\\
    QSTAR & surface moisture scale  & $\frac{kg}{kg}$ \\
    SPEED & surface wind speed  & $\frac{m}{s^2}$ \\
    TAUX & eastward surface stress & $\frac{N}{m^2}$ \\
    TAUY & northward surface stress & $\frac{N}{m^2}$ \\
    TLML & surface air temperature & $K$ \\
    ULML & surface eastward wind & $\frac{m}{s}$ \\
    VLML & surface northward wind &$\frac{m}{s}$ \\
    \end{tabular}
\end{table*}

\begin{figure*}[ht]
     \centering
     \begin{subfigure}[b]{\textwidth}
         \centering
         \includegraphics[trim={0mm 4.5mm 0mm 0mm},clip,scale=0.4]{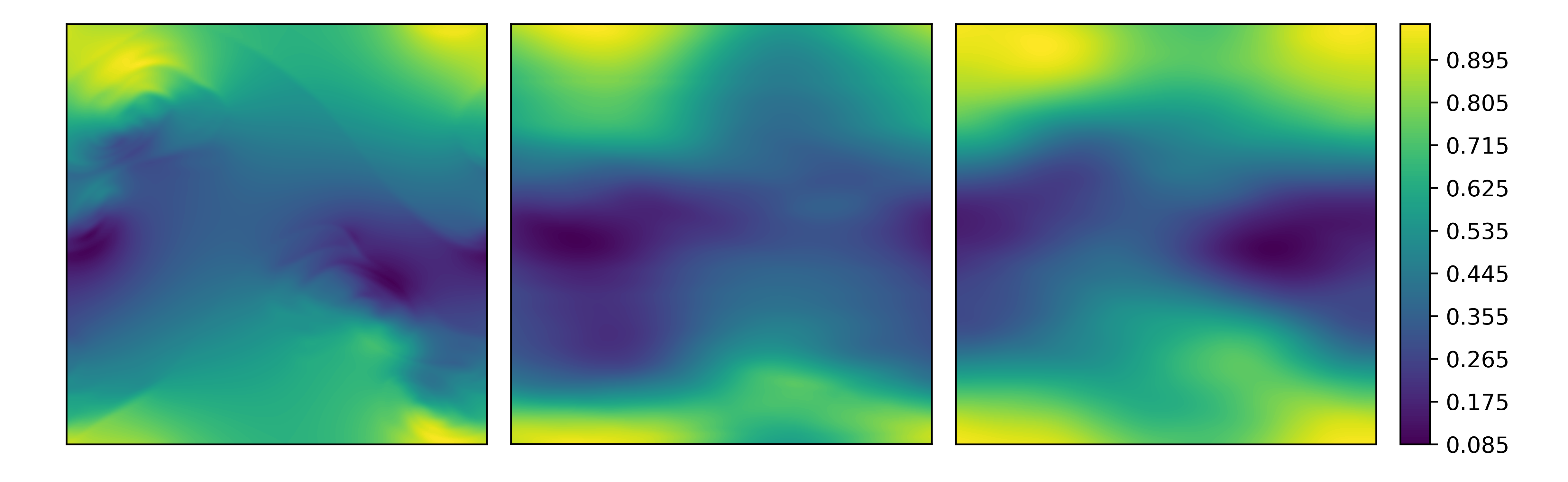}
         \caption{Horizontal velocity for the Shear Layer experiment}
         \label{fig:shear layer predictions}
         \end{subfigure}
     \begin{subfigure}[b]{\textwidth}
         \centering
         \includegraphics[trim={0mm 4.5mm 0mm -2mm},clip,scale=0.4]{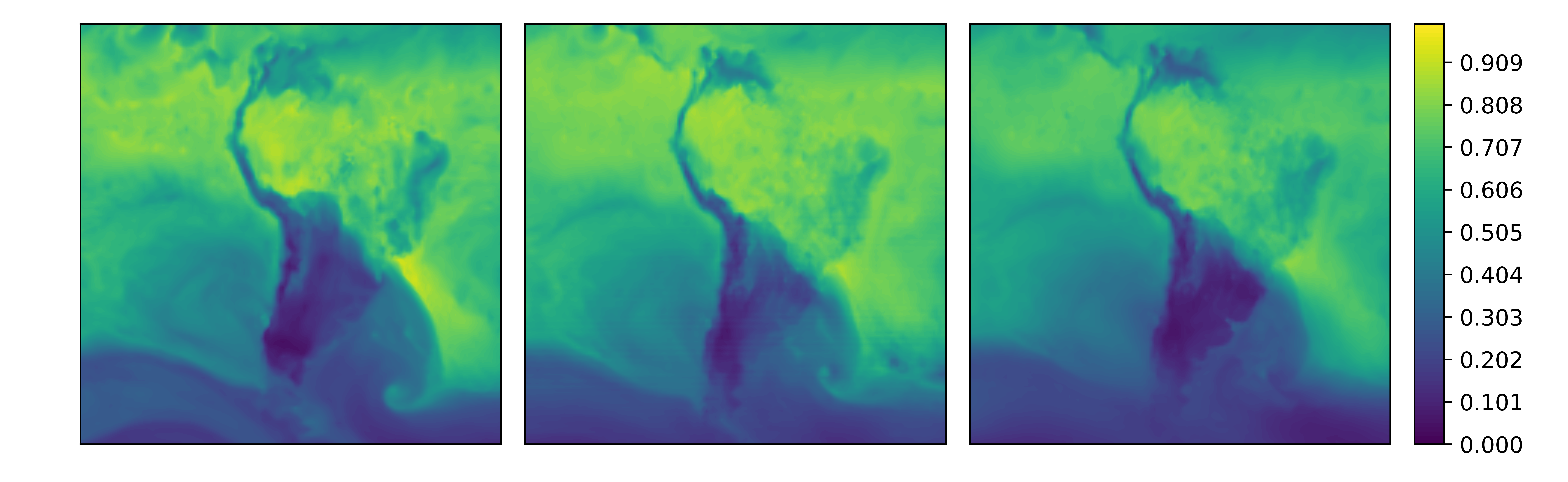}
    \caption{Surface-level specific humidity over South America.    \label{fig:humidity predictions}}
     \end{subfigure}
     \begin{subfigure}[b]{\textwidth}
         \centering
         \includegraphics[trim={0mm 3.3mm 0mm -3mm},clip,scale=0.4]{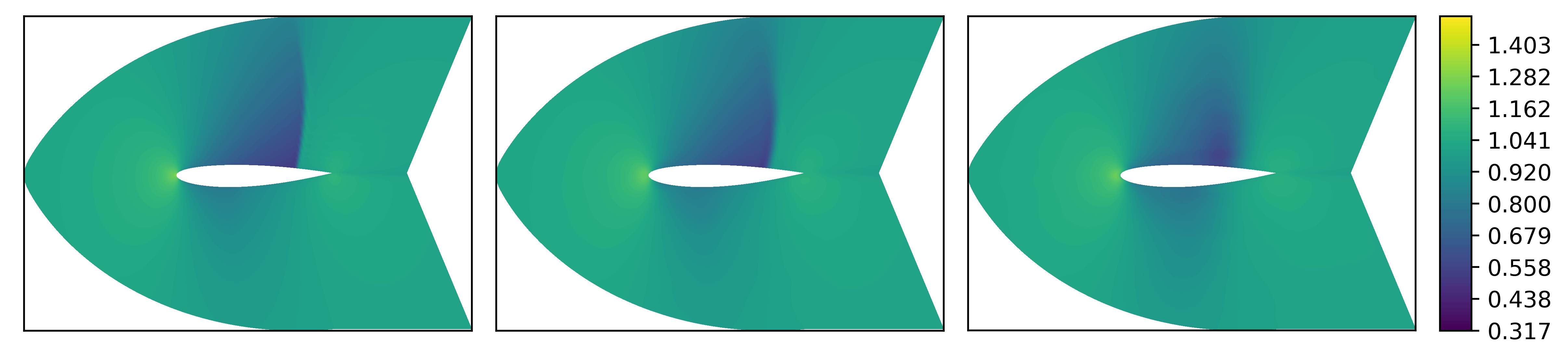}
    \caption{Pressure distributions around an airfoil.}
    \label{fig:airfoil predictions}
     \end{subfigure}
        \caption{These figures display examples of the ground truth, the target which the FNO-DSE, FNO, or Geo-FNO attempt to match. Left: Ground Truth. Center: FNO-DSE Right: FNO for (a) and (b) and Geo-FNO for (c).}
        \label{fig:predictions}
\end{figure*}

\begin{table*}[h]
    \centering
    \caption{The mean and standard deviation \emph{of the median test errors} among 10 models trained with different random seeds at initialization of the structured proposed method for the \emph{Burgers'} experiments.}
    \begin{tabular}{l l c c}
    \toprule
        \textbf{Data Distribution} & Model & Method & \textbf{Mean}$\pm$\textbf{Std}\\
        \midrule
        {\it Uniform} & FNO & FFT & 0.0650 $\pm$ 0.0051\% \\
        {\it Uniform} & FNO & DSE & 0.0653 $\pm$ 0.0074\% \\
        {\it Contracting-Expanding} & FNO & FFT+Interpolation & 0.2076 $\pm$ 0.0087\% \\
        {\it Contracting-Expanding} & FNO & DSE & 0.1973 $\pm$ 0.0069\% \\
    \end{tabular}
    
    \label{tab:burgers initializations}
\end{table*}

\begin{figure*}[h!]
  \centering
  
  \begin{subfigure}{0.5\textwidth}
    \centering
    \includegraphics[width=\linewidth]{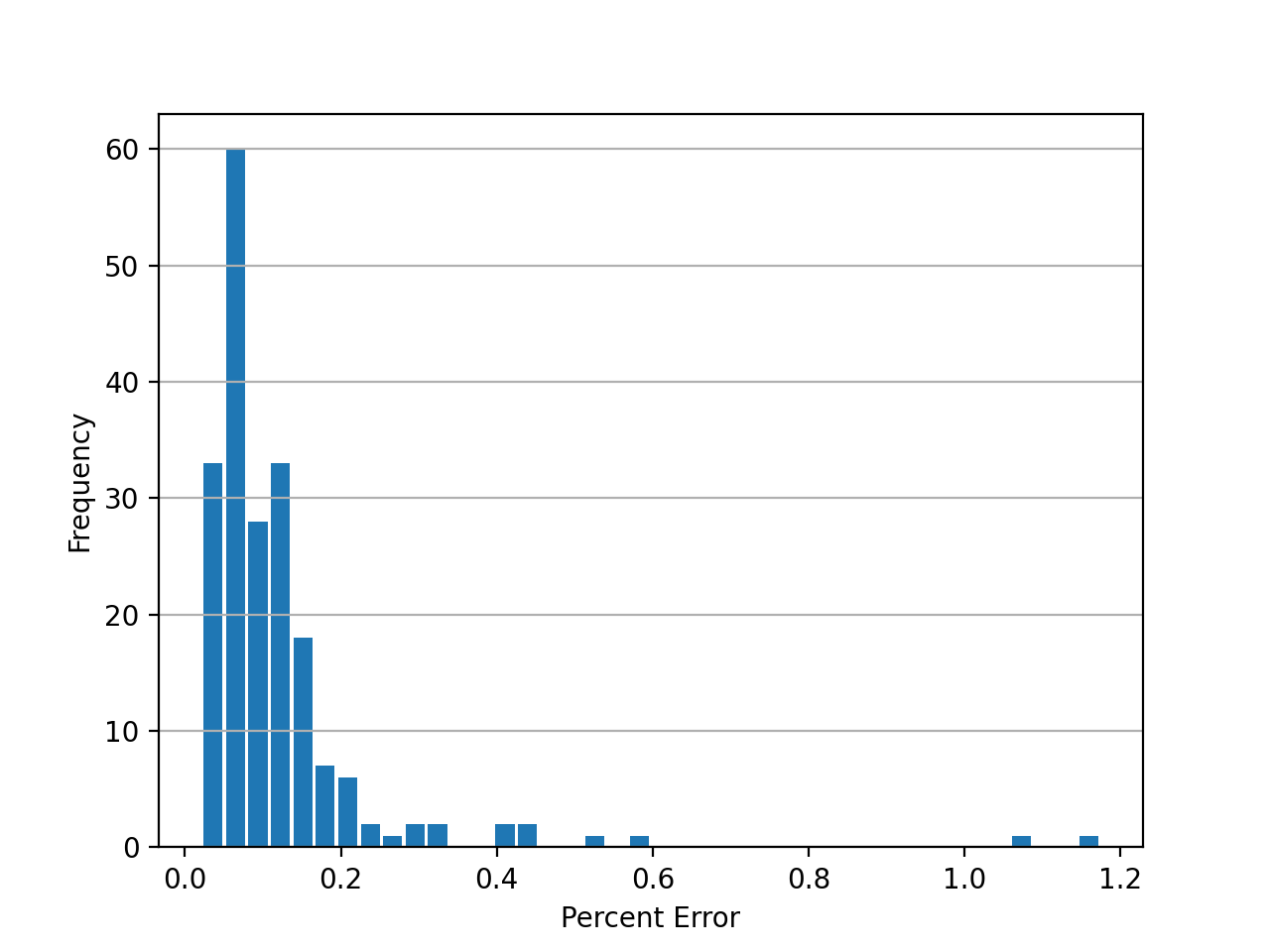}
    \caption{\emph{Uniform}.}
    \label{fig:burgers uniform histogram}
  \end{subfigure}%
  \begin{subfigure}{0.5\textwidth}
    \centering
    \includegraphics[width=\linewidth]{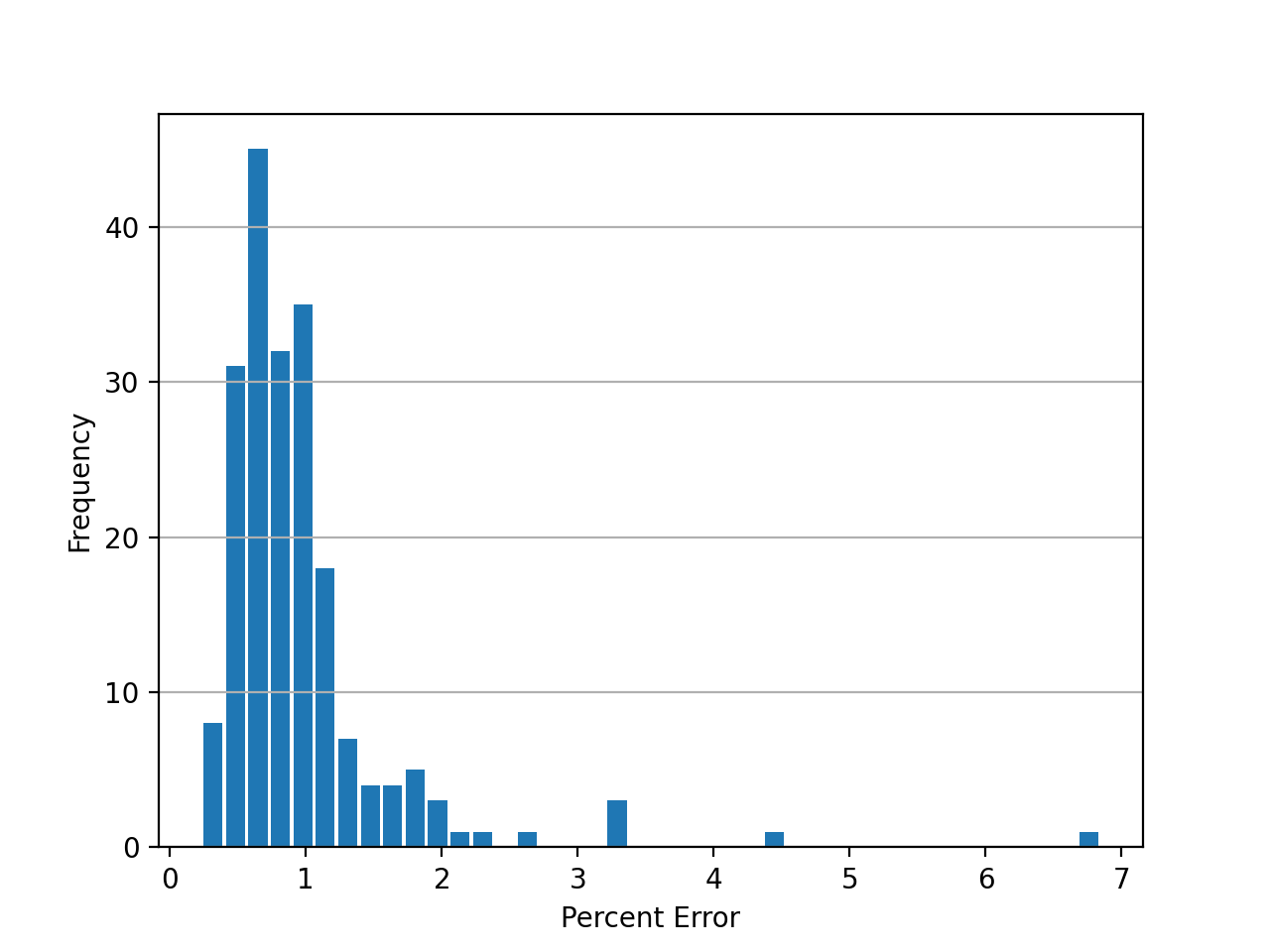}
    \caption{\emph{Contracting-Expanding}.}
    \label{fig:burgers conexp histogram}
  \end{subfigure}%
  
  \caption{Histograms displaying the distribution of test errors for the different point distributions of the \emph{Burgers'} numerical experiments using the proposed method.}
  \label{fig:burgers histograms}
\end{figure*}

\begin{figure*}[h!]
  \centering
  \begin{subfigure}{0.5\textwidth}
    \centering
    \includegraphics[width=\linewidth]{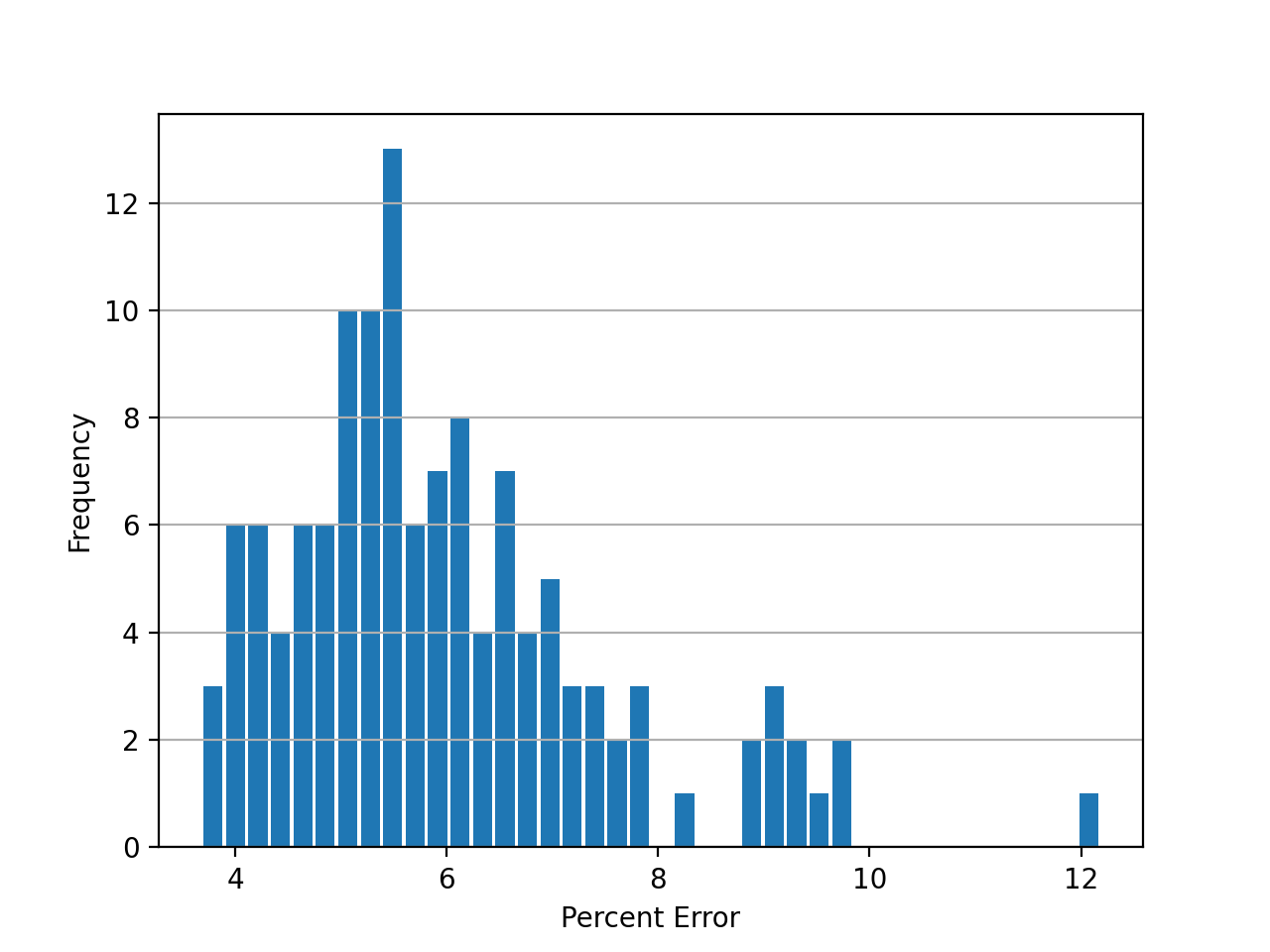}
    \caption{\emph{Shear Layer}.}
    \label{fig:shear layer histogram}
  \end{subfigure}%
  \begin{subfigure}{0.5\textwidth}
    \centering
    \includegraphics[width=\linewidth]{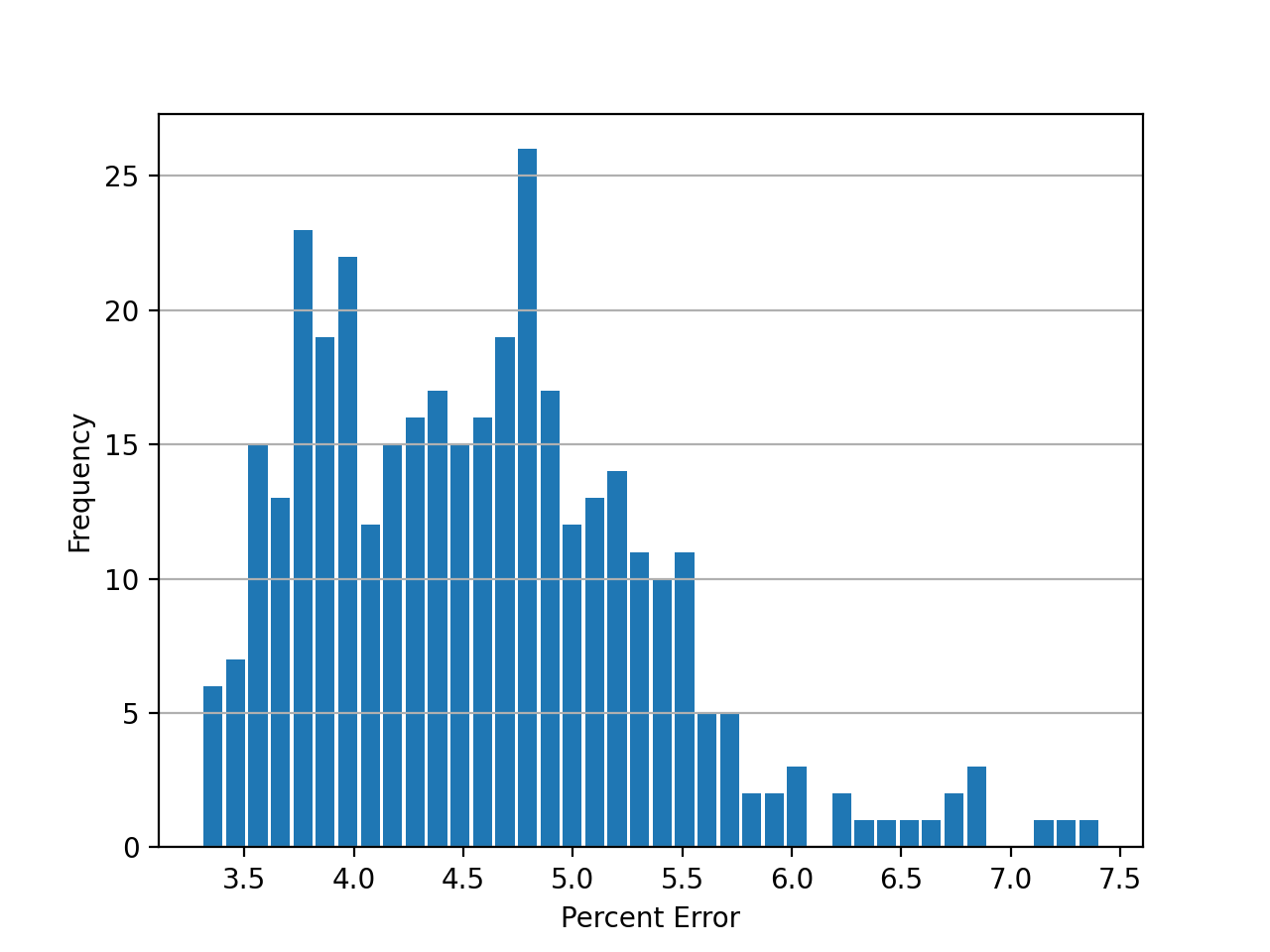}
    \caption{\emph{Surface-Level Specific Humidity}.}
    \label{fig:humidity histogram}
  \end{subfigure}
  \caption{Histograms displaying the distribution of test errors for the two-dimensional numerical experiments on the lattice using the proposed method with the FNO model.}
  \label{fig:2d histograms}
\end{figure*}

\begin{table*}
  \centering
  \caption{Average training times for several experiments as the number of modes is varied.}
  \begin{tabular}{c c c c c c c c c}
    \toprule
    Modes & \multicolumn{2}{c}{Burgers} & \multicolumn{2}{c}{Humidity} & \multicolumn{2}{c}{Airfoil} & \multicolumn{2}{c}{SWE} \\
     & FFT & DSE & FFT & DSE & Geo-FNO & Direct & SFNO & DSE \\
    \midrule
    8  & 0.93 & 0.81 & 15.3 & 12.3 & 3.5 & 1.9 & 72.6 & 17.1 \\
    10 & 0.94 & 0.81 & 15.3 & 12.1 & 4.4 & 2.5 & 72.3 & 17.1 \\
    12 & 0.94 & 0.81 & 15.0 & 12.6 & 5.6 & 3.3 & 72.0 & 17.6 \\
    14 & 0.94 & 0.81 & 15.2 & 12.6 & 6.9 & 4.0 & 72.1 & 17.3 \\
    16 & 0.95 & 0.82 & 15.2 & 12.6 & 8.5 & 5.2 & 72.1 & 17.3 \\
    18 & 0.95 & 0.82 & 15.2 & 12.6 & 10.4 & 7.0 & 72.3 & 18.1 \\
    20 & 0.95 & 0.82 & 15.4 & 12.8 & 12.5 & 9.1 & 72.4 & 17.4 \\
    22 & 0.95 & 0.82 & 15.5 & 13.1 & 14.8 & 10.3 & 72.8 & 18.6 \\
    24 & 0.95 & 0.83 & 15.3 & 12.9 & 17.2 & 11.8 & 72.1 & 17.8 \\
    26 & 0.95 & 0.83 & 15.2 & 13.2 & 20.1 & 14.1 & 72.6 & 18.1 \\
    28 & 0.94 & 0.83 & 15.0 & 12.9 & -- & 16.0 & 73.0 & 19.5 \\
    30 & 0.94 & 0.83 & 15.4 & 13.1 & -- & 18.0 & 72.5 & 18.5 \\
    32 & 0.94 & 0.83 & 15.2 & 13.1 & -- & 20.7 & 72.3 & 20.71 \\
  \end{tabular}
  \label{tab:ablation studies}
\end{table*}

\begin{table*}[htbp]
    \centering
    \caption{Model sizes in terms of the number of parameters. Within an experiment, the models are constructed to have similarly sized convolution and spectral convolution layers. This results in the FFNO often having an order of magnitude fewer parameters, a key contribution outlined by \cite{tran2023factorized}. The UFNO models have the most parameters due to the extra U-Net layers, and the Geometric Layer results in a slightly higher number of parameters as well.}
    \label{tab:models sizes}
    \begin{tabular}{l l c c}
        \toprule
        \textbf{Model} & \textbf{Method}  & {\textbf{No. Parameters}} & {\textbf{No. Modes}}\\
        \midrule
        \multicolumn{3}{l}{\textbf{1D: Burgers' Equation}} \\
        \multicolumn{3}{l}{\textit{Equispaced Distribution:}} \\
        \multirow{2}{*}{FNO} & DSE & 287425 & 16\\
        & FFT & 287425 & 16\\
        \multicolumn{3}{l}{\textit{Contracting-Expanding Distribution:}} \\
        \multirow{2}{*}{FNO} & DSE & 287425 & 16\\
        & Interpolation & 287425 & 16\\
        \midrule
        \multicolumn{3}{l}{\textbf{2D Lattice: Shear Layer}} \\
        \multirow{2}{*}{FNO} & DSE & 3162881 & 20\\
        & Full Grid & 3162881 & 20 \\
        \multirow{2}{*}{UFNO} & DSE & 3750401 & 20\\
        & Full Grid & 3750401 & 20 \\
        \multirow{2}{*}{FFNO} & DSE & 423041 & 20\\
        & Full Grid & 423041 & 20\\
        \midrule
        \multicolumn{3}{l}{\textbf{2D Lattice: Specific Humidity}} \\
        \multirow{2}{*}{FNO} & DSE & 8398049 & 32\\
        & Full Grid & 8398049 & 32 \\
        \multirow{2}{*}{UFNO} & DSE & 8691553 & 32\\
        & Full Grid & 8691553 & 32 \\
        \multirow{2}{*}{FFNO} & DSE & 537697 & 32\\
        & Full Grid & 537697 & 32 \\
        \midrule
        \multicolumn{3}{l}{\textbf{2D Point Cloud: Flow Past Airfoil}} \\
        \multirow{2}{*}{FNO} & DSE & 1188577 & 12\\
        & Geometric Layer & 1250339 & 12 \\
        \multirow{2}{*}{UFNO} & DSE & 1482081 & 12\\
        & Geometric Layer & 1840163 & 12\\
        \multirow{2}{*}{FFNO} & DSE & 500673 & 14\\
        & Geometric Layer & 526787 & 14\\
        \midrule
        \multicolumn{3}{l}{\textbf{2D Point Cloud: Elasticity}} \\
        \multirow{2}{*}{FNO} & DSE & 1484289 & 12 \\
        & Geometric Layer & 1546403 & 12 \\
        \multirow{2}{*}{UFNO} & DSE & 1778049 & 12 \\
        & Geometric Layer & 1840163 & 12 \\
        \multirow{2}{*}{FFNO} & DSE & 526787 & 14 \\
        & Geometric Layer & 533441 & 14 \\
        \bottomrule
        \multicolumn{2}{l}{\textbf{Random Spherical Point Cloud: Shallow Water Equations}} \\
        \multirow{3}{*}{SFNO} & DSE & 39749251 & $l=22$ \\
        & Gaussian Interpolation & 39201536  & $l=22$\\
        & Linear Interpolation & 39201536  & $l=22$\\
        \multirow{3}{*}{FNO} & DSE & 31978563 & 20 \\
        & Gaussian Interpolation & 39201536 & 20 \\
        & Linear Interpolation & 39201536 & 20 \\
        \bottomrule
    \end{tabular}%
\end{table*}

\subsection{Generalization Capabilities for the SWE Experiments.}

First, we provide the errors of the FNO and SFNO on their respective grids for the Shallow Water Equations experiments on the Sphere in Table \ref{tab:sfno original grid results}. We find that the training times are approximately 30 seconds, twice that of the DSE approaches on the unstructured data, and the errors over the collocation points are equivalent to those of the DSE approach.

We also present the results for the SFNO-DSE on new sets of points which are unseen at training time. We vary the number of points, providing a sensitivity analysis. The results, in Table \ref{tab:sfno smm varying grids results}, show that the error remains constant.

\begin{table*}[]
    \centering
    \begin{tabular}{c c c }
        \toprule
        Method & Training Time (per epoch) & Error over Collocation Points \\
        \midrule
        SFNO & 30s & 3.60\%  \\
        FNO & 32s & 5.74\%  \\
    \end{tabular}
    \caption{Accuracy of the SFNO and FNO for the Spherical Shallow Water Equations when trained on the original $256 \times 512$ grid which has been sub-sampled to a $51 \times 102$ grid. At inference time, we calculate the error over the collocation points, i.e. those used in the random distribution. These models achieve similar results to the proposed approach when the data has not undergone interpolation.}
    \label{tab:sfno original grid results}
\end{table*}

\begin{table*}[]
    \centering
    \begin{tabular}{c c c}
        \toprule
        Approximate No. Points & Error  \\
        \midrule
        4000 & 6.36\%   \\
        8000 & 5.64\%   \\
        15000 & 5.44\%  \\
        29000 & 5.39\%  \\
        50000 & 5.43\%  \\
    \end{tabular}
    \caption{Accuracy of the SFNO-DSE on test sets of the Spherical Shallow Water Equations experiment with new random distributions of varying size. Each test set consists of 64 samples. The model was trained on data sets of approximately 5000 points. All point distributions and test samples used in this evaluation were not seen during training. The results show that the proposed method is able to generalize to new point distributions of varying resolution. }
    \label{tab:sfno smm varying grids results}
\end{table*}

\end{document}